\documentclass[10pt]{article} %
\usepackage[preprint]{tmlr-style-file-main/tmlr}

\usepackage[utf8]{inputenc}
\usepackage[T1]{fontenc}

\usepackage{comment}
\usepackage{amsmath}
\usepackage{amsfonts}
\usepackage{amssymb}

\usepackage{authblk}

\usepackage{graphicx}

\usepackage{adjustbox}

\usepackage{grffile}
\usepackage{longtable}
\usepackage{wrapfig}
\usepackage{rotating}
\usepackage[normalem]{ulem}
\usepackage{textcomp}
\usepackage{capt-of}
\usepackage{hyperref}
\hypersetup{breaklinks=true}
\usepackage{url}

\usepackage[title]{appendix}

\usepackage{algorithmic}
\usepackage{graphicx}
\usepackage{xcolor}
\usepackage{multirow} 
\usepackage{xcolor}
\usepackage[ruled,linesnumbered,boxed]{algorithm2e}
\usepackage[inline]{enumitem}
\usepackage{booktabs}
\usepackage[skip=0pt]{caption}
\usepackage{acronym}
\usepackage{url}

\usepackage{xcolor}

\usepackage{amsthm}
\usepackage{dsfont}
\usepackage{color}

\usepackage{float}

\usepackage{subcaption}

\usepackage{siunitx}%

\usepackage{lipsum}

\usepackage{listings}
\usepackage{xcolor}

\definecolor{codegreen}{rgb}{0,0.6,0}
\definecolor{codegray}{rgb}{0.5,0.5,0.5}
\definecolor{codepurple}{rgb}{0.58,0,0.82}
\definecolor{backcolour}{rgb}{0.95,0.95,0.92}

\lstdefinestyle{mystyle}{
    backgroundcolor=\color{backcolour},   
    commentstyle=\color{codegreen},
    keywordstyle=\color{magenta},
    numberstyle=\tiny\color{codegray},
    stringstyle=\color{codepurple},
    basicstyle=\ttfamily\footnotesize,
    breakatwhitespace=false,         
    breaklines=true,                 
    captionpos=b,                    
    keepspaces=true,                 
    numbers=left,                    
    numbersep=5pt,                  
    showspaces=false,                
    showstringspaces=false,
    showtabs=false,                  
    tabsize=2
  }

\newtheorem{property}{Property}

\newtheorem*{definition*}{Definition}

\hypersetup{
 pdfauthor={Gabriel Bénédict},
 pdftitle={},
 pdfkeywords={},
 pdfsubject={},
 pdfcreator={Emacs 26.1 (Org mode 9.1.14)},
 pdflang={English}}

\newcommand{\header}[1]{\par\noindent\textbf{#1}}

\usepackage{amsmath,amsfonts,bm}

\def\eqref#1{equation~\ref{#1}}

\def\1{\bm{1}}

\DeclareMathAlphabet{\mathsfit}{\encodingdefault}{\sfdefault}{m}{sl}
\SetMathAlphabet{\mathsfit}{bold}{\encodingdefault}{\sfdefault}{bx}{n}

\title{sigmoidF1: A Smooth F1 Score Surrogate Loss for Multi\-label Classification}

\author{\name Gabriel Bénédict \email g.benedict@uva.nl \vspace{-\baselineskip} \\
        \addr RTL Nederland B.V. \& University of Amsterdam 
        \AND
        \name Hendrik Vincent Koops \email vincent.koops@rtl.nl\\ %
        \addr RTL Nederland B.V.
        \AND
        \name Daan Odijk \email daan.odijk@rtl.nl \\
        \addr RTL Nederland B.V.        
        \AND
        \name Maarten de Rijke \email m.derijke@uva.nl \\
        \addr University of Amsterdam
        }

\def\month{09}  %
\def\year{2022} %
\def\openreview{\url{https://openreview.net/forum?id=gvSHaaD2wQ}} %

\begin{document}

\maketitle

\begin{abstract}
\noindent%
Multilabel classification is the task of attributing multiple labels to examples via predictions. 
Current models formulate a reduction of the multilabel setting into either multiple binary classifications or multiclass classification, allowing for the use of existing loss functions (sigmoid, cross-entropy, logistic, etc.). 
These multilabel classification reductions do not accommodate for the prediction of varying numbers of labels per example. Moreover, the loss functions are distant estimates of the performance metrics. 
We propose \emph{sigmoidF1}, a loss function that is an approximation of the macro F1 score that 
\begin{enumerate*}[label=(\roman*)]
\item is smooth and tractable for stochastic gradient descent at training time, 
\item naturally approximates a multilabel metric, and 
\item estimates both label suitability and label counts. 
\end{enumerate*}
We show that any confusion matrix metric can be formulated with a smooth surrogate. 
We evaluate the proposed loss function on text and image datasets, and with a variety of metrics, to account for the complexity of multilabel classification evaluation. 
sigmoidF1 outperforms other loss functions on one text and three image datasets over several metrics. 
These results show the effectiveness of using inference-time metrics as loss functions for non-trivial classification problems like multilabel classification. 

\end{abstract}

\section{Introduction}
\label{sec:org662677c}

Many real-world classification problems are challenging because of unclear (or overlapping) class-boundaries, subjectivity issues, and disagreement between annotators.

\begin{figure*}[!t]
\centering
         \centering

         \includegraphics[width=0.9\linewidth]{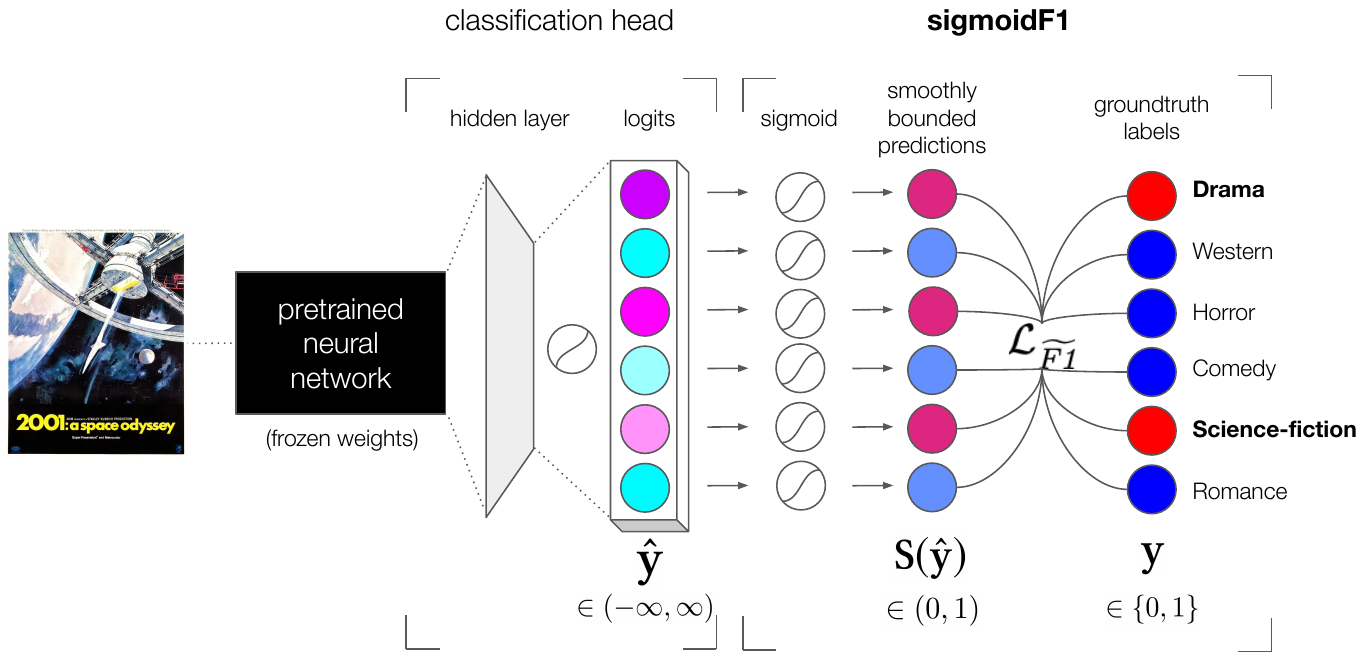}
         \caption{Experimental setup for \emph{sigmoidF1} as a loss function for \textbf{multilabel classification}. Here, a movie poster image is fed to a pre-trained network with a custom classification head that outputs logits (i.e., unbounded values) for each class (i.e., movie genre). At training time, a sigmoid function forces logits towards either $-1$ or $1$, respectively negative and positive predictions (illustrated by the darker colors). Confusion matrix metrics and macro F1 can then subsequently be computed. Here, $S(\hat{y}_{horror})$ is close to 1, but the ground truth data claims that \emph{2001: a space odyssey} is not a horror movie; this approximately corresponds to a false positive. Note that $\mathcal{L}_{\widetilde{\mathit{F1}}}$ is computed over a whole batch at training time as a macro measure with the formulas in Sections~\ref{sec:smoothConf} and~\ref{sec:orgc5d29d7}. With this setup, one can optimize directly for the metric of interest at training time. Our image and text classification tasks below show improved results when compared to existing losses.}
\label{fig:architecture}
\end{figure*}

Multilabel learning tasks are common, e.g., document and text classification often deal with multilabel problems~\citep{IRClassStat, textCategorization, statTextCategorization, documentClassification}, as do query classification~\citep{queryClassification, introIR}, image classification~\citep{imageClassification, faceDetection} and product classification~\citep{Amoualian2020SIGIR2E}.
Existing optimization frameworks typically split the task into known problems and sum over existing losses $\sum \mathcal{L}_{\text{MC}}$, with $\mathcal{L}_{\text{MC}}$ any multiclass classification loss – oftentimes variations of the cross-entropy or logistic loss. \citet{OVATheory} define these frameworks as \emph{multilabel reduction} techniques; \citet{multilabelReduction} put an emphasis on two: One-Versus-All (OVA)\footnote{This was already described in~\citep{OVA2} and further formalized in~\citep{OVATheory}.} and Pick-All-Labels (PAL). For example, if $C$ is the number of possible classes, OVA and PAL reformulate the multilabel problem to $C$ binary classification and $C$ multiclass classification problems, respectively (see Section~\ref{section:background:estimate}). These methods assume that, for one example, label probabilities (a.k.a.\ Bayes Optimal Classifier~\citep{OVA2}) are marginally independent of other label probabilities. \citeauthor{multilabelReduction} show mathematically and empirically that reduction methods (OVA and PAL) can optimize for precision or recall, but not for both precision and recall at once. More generally, a shortcoming shared by OVA and PAL is their reliance on the binary or multiclass classification setting and the lack of a pure multilabel approach – inspired by binary classification literature (see most recently \citep{multiclassOptDL2} and their F1 surrogate loss functions on 3-layer neural networks). 
We are not aware of a metric surrogate loss function that deals with multilabel classification in a modern deep learning setting in a single task. Figure~\ref{fig:architecture} illustrates our approach with a concrete example of classifying a movie poster into movie genres with a single loss function: \emph{sigmoidF1}.

\header{Proposed solution to multilabel problems.}
We propose a loss function $\mathcal{L}_{\widetilde{\mathit{F1}}}$ that
\begin{enumerate*}[label=(\roman*)]
\item naturally approximates the macro $F1$ classification metric (see Table~\ref{table:overallresults}),
\item estimates label probabilities and label counts (see Eq.~\ref{eq:unboundedF1}), 
and 
\item is decomposable for stochastic gradient descent at training time (see Section~\ref{subsection:properties} and Figure~\ref{fig:sigmoid}). 
\end{enumerate*}
Our proposed solution is to minimize a surrogate of the F1 metric as a loss. Strictly speaking, we minimize $1-\widetilde{\mathit{F1}}$, where $\widetilde{\mathit{F1}}$ is a smooth version of $F1$. Using a metric as a loss function is unpopular for metrics that require a form of thresholding (e.g., counting the number of true positives), as minimizing a step loss function (a.k.a. 0-1 loss) is intractable. The soft margin for support vector machines is an early example, where the intractability of the direct 0-1 loss optimization is overcome with the hinge loss~\citep{softMargin}. 
We resolve this by approximating the step function by a sigmoid curve (see Figure \ref{fig:architecture}).

\header{Main contributions.}
We introduce \emph{sigmoidF1}, an F1 score surrogate, with a sigmoid function acting as a surrogate thresholding step function.
\emph{sigmoidF1} allows for the use of the F1 metric that simultaneously optimizes for label prediction and label counts in a single task.
\emph{sigmoidF1} is benchmarked against loss functions commonly used in multilabel learning and other existing multilabel models. We show that our custom losses improve predictions over current solutions on several different metrics, across text and image classification tasks. PyTorch and TensorFlow source code are made available.\footnote{\url{https://github.com/gabriben/metrics-as-losses}}

\section{Background}
\label{section:background}

We use a traditional statistical framework as a guideline for multilabel classification methods~\citep{tukey}. We distinguish the desired theoretical statistic (the \textbf{estimand}), its functional form (the \textbf{estimator}) and its approximation (the \textbf{estimate}); estimates can be benchmarked with \textbf{metrics}. We show how multilabel reduction estimators tend to reformulate the estimand and treat labels as marginally independent. For example, by treating a multilabel problem as a succession of binary classification tasks. However, with a proper estimator, it is possible to directly model the estimand. If F1 score is indeed the statistic of interest (i.e. estimand), our proposed loss function, \emph{sigmoidF1}, accommodates for the true estimand.

We define a learning algorithm $\mathcal{F}$ (i.e., a class of estimators) that maps inputs to outputs given a set of hyperparameters \(\mathcal{F}(\cdot ; \Theta): \mathcal{X} \rightarrow \mathcal{Y}\). We consider a particular case, with the input vector \(\mathbf{x} = \{x_1, \ldots, x_n\}\) and each observation is assigned $k$ labels (one or more) \(\mathbf{l} = \{l_1, \ldots, l_C\}\) out of a set of $C$ classes. \(y_{i}^{j}\) are binary variables, indicating presence of a label for each observation \(i\) and class \(j\). Together, they form the matrix output $\mathbf{Y}$. This is our multilabel setting. Note that multiclass classification can be considered as an instance of multilabel classification, where a single label is attributed to an example.

\subsection{Estimand and definition of the risk}
\label{section:background:estimand}

We distinguish between two scenarios: the \emph{multiclass} and the \emph{multilabel} scenario. 
In the multiclass scenario, a single example is attributed one class label (e.g., classification of an animal on a picture). 
In the multilabel scenario, a single example can be assigned more than one class label (e.g., movie genres). 
We focus on the latter. 
For a particular set of inputs $\mathbf{x}$ (e.g., movie posters) and outputs $\mathbf{Y}$ (e.g., movie genre(s)), the risk formulation is the same as in~\citep{multilabelReduction}:
\begin{equation}
R_{\mathrm{ML}}(\mathcal{F}) = \mathbb{E}_{(\mathbf{x}, \mathbf{Y})}\left[\mathcal{L}_{\mathrm{ML}}(\mathbf{Y}, \mathcal{F}(\mathbf{x}))\right].
\end{equation}
The learning algorithm $\mathcal{F}$ is the estimand, the theoretical statistic. For one item $x_i$, the theoretical risk defines how close the estimand can get to that deterministic output vector $\mathbf{y}_{i}$.
In practice, statistical models do output probabilities $\mathbf{\hat{y}}_{i}$ via an estimator and its estimate (also called propensities or suitabilities~\citep{multilabelReduction}). The solution to that stochastic-deterministic incompatibility is either to convert the estimator to a deterministic measure via decision thresholds (e.g., traditional cross-entropy loss), or to treat the estimand as a stochastic measure (our \emph{sigmoidF1} loss proposal).

\subsection{Estimator: the functional form}
\label{section:background:estimator}

The estimator $f \in \mathcal{F}$ is any minimizer of the risk $R_{\mathrm{ML}}$. Predicting multiple labels per example comes with the assumption that labels are non mutually-exclusive.

\begin{definition*}
\label{prop:estimator}
  The multilabel estimator of $y_{i}^{j}$ is dependent on the input and other ground truth labels for that example,  $\hat{y}_i^j = f(x, y_{i}^{1}, \ldots, y_{i}^{j-1}) = P(y_i^j = 1 | y_{i}^{1}, \ldots, y_{i}^{j-1}, x_i)$.
\end{definition*}

By proposing this general formulation, we entrench that mutually-inclusive characteristic in the estimator. Contrary to \citet{multilabelReduction}, our definition above models interdependence between labels and deals with thresholding for the estimate at training time for free. \citet{F-inference} show that an estimator of an F-score can be used at inference time for multilabel classification, when using probabilistic models where parameter estimation is possible (e.g., decision trees, probabilistic classifier chains). When it is not possible, we resort to defining a loss function.

\subsection{Estimate: approximation via a loss function}
\label{section:background:estimate}

Most of the literature on multilabel classification can be characterized as multilabel reductions~\citep{multilabelReduction}: an approximation of the original multilabel problem via a loss function $\mathcal{L}(\mathbf{y}_i, f)$. It can take different forms.

\header{One-versus-all (OVA)} is a reformulation of the multilabel classification task to a sequence of $C$ binary classifications ($f^1, \ldots, f^C$), with $C$ the number of classes, $\mathcal{L}_{\mathrm{OVA}}(\mathbf{y}_i, f) = \sum_{c = 1}^C \mathcal{L}_{\mathrm{BC}}\left(y_i^{c}, f^{c}\right)$
where $\mathcal{L}_{\mathrm{BC}}$ is a binary classification loss (binary relevance \citep{OVA1, hammingLoss, OVA2}), most often logistic loss.  
Minimizing binary cross-entropy is equivalent to maximizing for log-likelihood~\cite[\S4.3.4]{Bishop}.

\header{Pick-all-labels (PAL)} gives the loss function $\mathcal{L}_{\mathrm{PAL}}(\mathbf{y}_i, f) = \sum_{c = 1}^C y_{i}^c \cdot \mathcal{L}_{\mathrm{MC}}(y_i^c, f)$,
with $\mathcal{L}_{\mathrm{MC}}$ a multiclass loss (e.g., softmax cross-entropy). In this formulation, each example $(x_i, \mathbf{y}_i)$ is converted to a multiclass framework, with one observation per positive label. The sum of inherently multiclass losses is used to represent the multilabel estimand. %

Multilabel reduction methods are characterized by their way of reformulating the estimand, the resulting estimator, and the estimate. This allows the use of existing losses: logistic loss (for binary classification formulations), sigmoid or softmax cross-entropy loss (for multiclass formulations). These reductions imply a reformulation of the estimator (a.k.a.\ Bayes Optimal) as follows:
\begin{equation}
  \hat{y}_i^j = f(x) = P(y_i^j = 1 | x_i).
\end{equation}
Contrary to our definition of the original multilabel estimator (Section~\ref{prop:estimator}), marginal independence of label propensities is assumed. In other words, the loss function becomes any monotone transformation of the marginal label probabilities $P(y_i^j = 1 | x)$~\citep{OVA2, multilabelMetrics, unifiedView}. In literature reviews, the multilabel reductions OVA and PAL have been coined as \emph{fit-data-to-algorithm}, as opposed to \emph{fit-algorithm-to-data} \citep{multilabelReview}, originally framed as \emph{problem transformation} and \emph{algorithm adaptation} respectively \citep{hammingLoss}). For the purpose of our narrative, we propose the following formalization of this dichotomy: \emph{fit-data-to-algorithm} formulates an additive loss over existing losses $\sum \mathcal{L}_{\mathcal{C}}$, with $\mathcal{L}_{\mathcal{C}}$ any classification loss and oftentimes a sum over all classes. This can be contrasted with \emph{fit-algorithm-to-data}, where a custom loss $\mathcal{L}^*$ is built for the multilabel task. We further discuss this in Section \ref{sec:org2aceb9f} and Table \ref{table:litrev}.

\subsection{Metrics: evaluation at inference time}
\label{section:background:metrics}

There is consensus on the usefulness of a confusion matrix and ranking metrics to evaluate multilabel classification models at inference time~\citep{multilabelMetrics, weightedMetrics, unifiedView}. Confusion matrix metrics come with caveats: most of these measures
\begin{enumerate*}[label=(\roman*)]
\item require hard thresholding, which makes them non-dif\-fer\-en\-tiable for stochastic gradient descent;
\item they are very sensitive to the number top labels to include $k$ \citep{decisionThreshold}; and 
\item they require aggregation choices to be made in terms of micro/macro/weighted metrics.
\end{enumerate*}
Common confusion matrix metrics are Precision, Recall, F1-score or one-error-loss; see~\citep{unifiedView} for others.

\subsection{Multilabel estimate: F1 metric as a loss}
\label{section:background:metricsAsLosses}

A model's out-of-sample accuracy is commonly measured on metrics such as AUROC, F1 score, etc. These reflect an objective catered towards evaluating the model over an entire ranking. Due to the lack of differentiability, these metrics cannot be directly used as loss functions at training time (in-sample).
\citet{optimizableLosses} propose a theoretical framework for deriving decomposable surrogates to some of these metrics. We propose our own decomposable surrogates tailored for multilabel classification (see Appendix~\ref{sec:evalMetrics}).

In a typical machine learning classification task, ground truth binary labels are compared to a probabilistic measure (or a reversible
transformation of a probabilistic measure such as a sigmoid or a softmax
function)~\citep{Bishop}. If the number $n_i$ of labels to be predicted per
example is known a priori, it is natural at training time to assign the $top_{n_i}$ predictions
to that example~\citep{lossTopKError, topKmulticlassSVM}. If the number of
labels per example is not known a priori, the question remains at both training and at inference time
as to how to decide on the number of labels to assign to each
example. This is generally done via a \emph{decision threshold}, that can be set globally for all
examples~\citep{threshForF1}. This threshold can optimize for specificity or
sensitivity~\citep{decisionThreshold} –  for per-class thresholding see~\cite{moviePosters}. 
In Section~\ref{section:method}, we propose an approach where this threshold is implicitly defined at training time, by using a loss function that penalizes explicitly for wrong label counts and fits to the original estimand in Definition~\ref{prop:estimator}: the F1 metric. In Section \ref{section:method}, we show how $F_1$ is formulated into a surrogate loss $\mathcal{L}_{\widetilde{\mathit{F1}}}$. Our contribution is thus in the continuation of the \emph{fit-algorithm-to-data} trend, because we propose a custom loss function.  That loss function is also the first to directly approximate the F1 score with non-divergent estimates (see Sections~\ref{subsection:properties} and \ref{subsec:unbounded} on \emph{boundedness}).

\section{Related Work}
\label{sec:org2aceb9f}

In Section~\ref{section:background:estimate}, we mentioned how existing solutions for multilabel tasks can be divided into \emph{fit-data-to-algorithm} solutions, which map multilabel problems to a known problem formulation like multiclass classification, and \emph{fit-algorithm-to-data} solutions, which adapt existing classification algorithms to the problem at hand~\citep{multilabelMethods}. In most of this work, the term \emph{multilabel classification} excludes \emph{extreme} (tens of thousands of labels) ~\citep[e.g.,][]{extremeClassification,millionsOfLabels, extremeMilliionsSlice}, \emph{hierarchical} (parent and children labels) ~\citep[e.g.,][]{lehmann2015dbpedia, XLNet, ULMFit} or \emph{multiclass} (single label per example) subfields. These subfields call for their own solutions, including label embeddings~\citep{extremeMultilabelEmbeddings} or negative mining~\citep{stochasticNegativeMining} for the \emph{extreme} usecase.

\header{Fit-data-to-algorithm.}
In fit-data-to-algorithm solutions, cross-entropy losses~\citep{Fisher, CE} are used at training time and thresholding is done at inference time to determine how many labels should be assigned to an instance. This has also been called multilabel reduction~\citep{multilabelReduction} and differs from multiclass-to-binary classifications~\citep{multiclassToBinary1, multiclassToBinary2, multiclassToBinary3}. We can further distinguish between One-versus-all (OVA) and Pick-all-labels (PAL) solutions~\citep{multilabelReduction} (see Section~\ref{section:background}).
In OVA, one reduces the classification problem to independent binary classifications~\citep{OVA1, hammingLoss, OVA2, OVATheory}. 
In PAL, one reformulates the task to independent multiclass classifications \citep{labelPowerset, extremeClassification, PAL}. 
The \textit{label powerset} approach considers each set of labels as a class~\citep{labelPowerset}. 
In Pick-One-Label (POL), a single multiclass example is created by randomly sampling a positive label \citep{PAL, extremeClassification}. 
Alternatively, \textit{ranking by pairwise comparison} is a solution where the dataset is duplicated for each possible label pair. Each duplicated dataset has therefore two classes and only contains instances that have at least one of the labels in the label pair. Different ranking methods exist~\citep{multilabelBackprop, pairwiseBinary, pairwiseNet}. Ranking loss has been shown to optimize for two Learning To Rank metrics~\citep{rankingNDCG}.
More recently, hierarchical datasets such as DBpedia~\citep{lehmann2015dbpedia} are used to fine-tune BERT-based models~\citep{XLNet, bigBird};  the latter publications use cross-entropy to predict the labels.

\begin{table}
\caption{\emph{SigmoidF1} and related loss formulations ordered by publication date. The solution column refers to our proposed formalization of the literature review on how to conduct multilabel classification: \emph{D2A} refers to \emph{fit-data-to-algorithm} (sum over existing or cross-entropy-like, CE-like, classification losses $\sum \mathcal{L}_{\mathcal{C}}$) and \emph{A2D} refers to \emph{fit-algorithm-to-data} (custom loss $\mathcal{L}^*$) }
\label{table:litrev}
\centering
\begin{adjustbox}{width=\columnwidth, center}
  \begin{tabular}{l llllll}    
  \toprule
  Method & Solution & Model type & Context & Implementation & Surrogated metric & Modality\\
    \midrule
  ACE [\citeauthor{Fisher}] & \emph{D2A} & Any & Any & CE-like & – & Any\\
  rankingLoss [\citeauthor{multilabelBackprop}] & \emph{D2A} & Any & Any & pair-rank & – & tabular\\    
  MFC [\citeauthor{sigmoid}] & – & Gaussian mixtures & Mispronunciation detection & sigmoid & $F_1$ & Text\\
  optLosses [\citeauthor{optimizableLosses}] & \emph{A2D} & Any & Any & – & $F_1$ & Theoretical\\
  focalLoss [\citeauthor{focalLoss}] & \emph{D2A} & Neural net & Imbalanced-multiclass & CE-like & – & Image\\
  deepF [\citeauthor{deepF}] & \emph{A2D} & Neural net & multilabel & CE-like & $F_1$ & Image\\
  softF1 [\citeauthor{softF1}] & \emph{A2D} & Neural net & Multilabel & Unbounded & $F_1$ & Image\\
  ASL [\citeauthor{ASL}] & \emph{D2A} & Neural net & Multilabel & CE-like & – & Image\\
  RS@k [\citeauthor{recallLoss}] & \emph{A2D} & Neural net & Similarity & sigmoid & Recall & Image\\
  polyLoss [\citeauthor{polyloss}] & \emph{A2D} & Neural net & Imbalanced-multiclass, \ldots & CE-like & – & Image\\
  sigmoidF1 [ours] & \emph{A2D} & Neural net & Multilabel & sigmoid & $F_1$ & Text \& Image\\
  \bottomrule
  \end{tabular}
  \end{adjustbox}
\end{table}

\header{Fit-algorithm-to-data.}
In fit-algorithm-to-data solutions, elements of the learning algorithm are changed (e.g., the back propagation procedure). Before focusing on the multilabel case, the multiclass literature has some examples of F1 surrogate loss functions in particular: in the context of SVMs, via pseudo linear functions~\citep{multiclassOpt1} or by learning a feasible confusion matrix~\citep{multiclassOpt1}; in the context of deep networks, by learning the surrogate loss function via a dedicated neural network in the binary classification case~\citep{binaryOpt}, by optimizing performance measures composed of true positive and true negative rates~\citep{multiclassOptDL} or via empirical utility maximization of F1
on 3-layer neural networks~\citep{multiclassOptDL2}. Early representatives of multilabel fit-algorithm-to-data solutions stem from heterogenous domains of machine learning. 
MultiLabel $k$-Nearest Neighbors \citep{ML-KNN}, MultiLabel Decision Tree~\citep{ML-DT}, Ranking Support Vector Machine (SVM) \citep{multilabelSVM} and backpropagation for multiLabel learning with a ranking loss~\citep{multilabelBackprop}. More recently, the idea of multi-task learning for \emph{label prediction} and \emph{label count prediction} was introduced \cite[ML\(_{\text{NET}}\),][]{multitaskLabel,multitaskLabelImages, tencent}. 
The literature has been clearly hinting at the usefulness of a single task loss function that approximates a metric.  A formulation similar to our loss \emph{unboundedF1} was proposed in an unpublished blog post, which was referred to as \emph{softF1}~\citep{softF1}. A similar proposal was to use the hinge loss as a decomposable surrogate for confusion matrix entries for binary classification~\citep{optimizableLosses}. Outside of the context of neural networks, the \emph{Maximum F1-score criterion} for automatic mispronunciation detection was proposed as an objective function to a Gaussian Mixture Model-hidden Markov model (GMM-HMM)~\citep{sigmoid}. A recent paper used recall as a loss function for image similarity~\citep{recallLoss}.
In parallel, there is a growing consensus that the original cross-entropy loss (\emph{fit-data-to-algorithm}) cannot solve all our problems. A variation of the cross-entropy loss adapted to multilabel classification has been proposed \citep{ASL, tencent}; it extends the multiclass sparse class representation setting~\citep{focalLoss, polyloss}.
In the ranking domain, LambdaLoss has been proposed to optimize directly for the lambdaRank metric~\citep{lambdaLoss}. In the theoretical space, \cite{optimizableLosses} have proposed a generic framework for decomposable metrics, including $F1$ as a theoretical fractional linear program. Table \ref{table:litrev} illustrates how \emph{sigmoidF1} differs from the methods listed in this paragraph.

An important limitation shared by existing \emph{fit-data-to-algorithm} and \emph{fit-algorithm-to-data} approaches is the lack of a unified loss framework that deals with multilabel classification and can approximate a metric of interest. \emph{sigmoidF1} computes an F1 surrogate loss over the aggregation of examples in a batch at training time.

\section{Method}
\label{section:method}

We introduce our approach for multilabel problems, with a smoothed confusion matrix metric as a loss (the original confusion matrix metrics rely on step functions and are therefore intractable, see for example the blue step function in Figure \ref{fig:sigmoid}). We first briefly define our learning setting and define the confusion matrix metrics in this setting more formally.

We use the binary classification setting (two classes) to simplify notation, without loss of generalization to the multilabel case.
In a typical binary classification problem with the label vector $\mathbf{y} = \{y_1 , \ldots, y_n\}$, predictions are probabilistic and it is necessary to define a threshold \(t\), at which a prediction is binarized. With \(\mathds{1}\) as an indicator function, \(\mathbf{y}^+ = \sum \mathds{1}_{\hat{\mathbf{y}} \geq t}\), \(\mathbf{y}^- = \sum \mathds{1}_{\hat{\mathbf{y}} < t}\) are thus the count of positive and negative predictions at threshold \(t\). Let \(\mathit{tp}\), \(\mathit{fp}\), \(\mathit{fn}\), \(\mathit{tn}\) be number of true positives, false positives, false negatives and true negatives respectively:
\begin{equation}
\label{eq:conf}
\begin{array}{ll}\mathit{tp} = \sum \mathds{1}_{\hat{\mathbf{y}} \geq t} \odot \mathbf{y}  & \mathit{fp} = \sum \mathds{1}_{\hat{\mathbf{y}} \geq t} \odot (\mathds{1} - \mathbf{y}) \\[.5em] \mathit{fn} = \sum \mathds{1}_{\hat{\mathbf{y}} < t} \odot \mathbf{y} & \mathit{tn} = \sum \mathds{1}_{\hat{\mathbf{y}} < t} \odot (\mathds{1} - \mathbf{y}),
\end{array}
\end{equation}
with \(\odot\) the component-wise multiplication sign.
For simplicity, in the formulation above and the ones that follow scores are calculated for a single class, therefore the sum is implicitly over all examples \(\sum_i\). This applies to the binary classification problem but also to our multilabel setting, when micro metrics are calculated (i.e., compute the metric value for each class, and then averaged over all classes). In the multilabel setting $\mathbf{y}$ can be substituted by $\mathbf{y}^j$ for each class $j$. Note that vectors could be trivially substituted by matrices ($\mathbf{Y}$) in Eq.~\ref{eq:conf} to obtain the macro formulation.
Given the four confusion matrix quadrants, we can generate further metrics like precision and recall (see Table \ref{tab:confusion-matrix} in Appendix~\ref{sec:evalMetrics}). 
However, none of these metrics are decomposable due to the hard thresholding, which is, in effect, a step function (see Figure \ref{fig:sigmoid}).

Next, we define desirable properties for decomposable thresholding, unbounded confusion matrix entries, and a sigmoid transformation that renders confusion matrix entries decomposable. Finally, we focus on a smooth F1 score.

\subsection{Desirable properties of decomposable thresholding}
\label{subsection:properties}

We define desirable properties for a decomposable sign function $f(u)$ as a surrogate of the above indicator function \(\mathds{1}_{\hat{\mathbf{y}} < t}\).

\begin{property}
  Boundedness: $|f(u)| < M$, where $M$ is an upper and lower bound.
\end{property}
The ground truth $\mathbf{y}$ is bounded between $[0,1]$ and thus it must be compared to a bounded prediction $\mathbf{\hat{y}}$, preferably bounded by $[0,1]$, to avoid further scaling.

\begin{property}
\label{prop:sat}
Saturation: $\int_{s}^{\infty} f^{-1}(u) = \int_{-\infty}^{-s} f(u) = \epsilon$, with $\epsilon$ a number close to zero and $s$ a saturation bound.
\end{property}
For the surrogate to be a proper sign function substitute, it is important to often return values close to 1 or 0. Saturation is defined in the context of neural network activation functions and refers to the propensity of iterative backpropagation to progressively lead to values very close to 0 or 1 after a long enough training period. Our aim is to reach that convergence quickly in order to force decisions towards 0 or 1 in order to be comparable to a step function. This highlights a tension: the sigmoid function should contrast outputs towards $0$ or $1$ but should not be too saturated, in order for the derivative at point $u$ to be non-null and information to flow back to the network~\citep{saturation}.

\begin{property}
  Dynamic Gradient: $f'(u) \gg 0 \quad \forall \; u \in [-s, s]$, where $s$ is the saturation bound.
\end{property}

\noindent%
Inside the saturation bounds $[-s, s]$, the derivative should be significantly higher than zero in order to facilitate stochastic gradient descent and backpropagation.
Note that the upper and lower limits of $f(u)$ are interchangeably $[-1,1]$ or $[0,1]$ in this paper and in the literature. The conditions above still apply after linear transformation.
Next, we show how our formalization of an unbounded F1 surrogate would not fulfill these properties and how our proposition of a smooth bounded alternative does.

\subsection{Unbounded confusion matrix entries}
\label{subsec:unbounded}

A first trivial remedy to allow for derivation of the sign function $f(u)$, is to define \emph{unbounded} confusion matrix entries by retaining the original logits (scores) when counting true positives, false negatives, etc. Countrary to the original confusion matrix definition in Eq.~\ref{eq:conf}, \(\overline{tp}\), \(\overline{fp}\), \(\overline{fn}\) and  \(\overline{tn}\) are not natural numbers anymore:
\begin{equation}
\label{eq:unbounded}
\mbox{}\hspace*{-2mm}\begin{array}{ll} \overline{\mathit{tp}} = \sum \hat{\mathbf{y}} \odot \mathbf{y}  & \overline{\mathit{fp}} = \sum \hat{\mathbf{y}} \odot (\mathds{1} - \mathbf{y}) \\[.5em] \overline{\mathit{fn}} = \sum (\mathds{1} - \hat{\mathbf{y}}) \odot \mathbf{y} & \overline{\mathit{tn}} = \sum (\mathds{1} - \hat{\mathbf{y}}) \odot (\mathds{1} - \mathbf{y}),
\end{array}
\end{equation}
where
\(\mathit{tp}\), \(\mathit{fp}\), \(\mathit{fn}\) and \(\mathit{tn}\) are now replaced by rough surrogates. The disadvantages are that the desirable properties mentioned above are not fulfilled, namely 
\begin{enumerate*}[label=(\roman*)]
\item \(\hat{\mathbf{y}}\) is unbounded and thus certain examples can have over-proportional effects on the loss; 
\item it is non-saturated; while non-saturation is desirable for activation functions~\citep{saturation}, here it would be desirable to tend towards saturation (i.e., tend to values close to 0 or 1, so as to give the most accurate predictions at any thresholding values at inference time); and 
\item the gradient of that linear function is 1 and therefore backpropagation will not learn depending on different inputs at this stage of the loss function.
\end{enumerate*}
However, this method has the advantage of resulting in a linear loss function that avoids the concept of thresholding altogether and is trivial to decompose for stochastic gradient descent.

\begin{figure}[tbp]
\centering
\includegraphics[width=\linewidth]{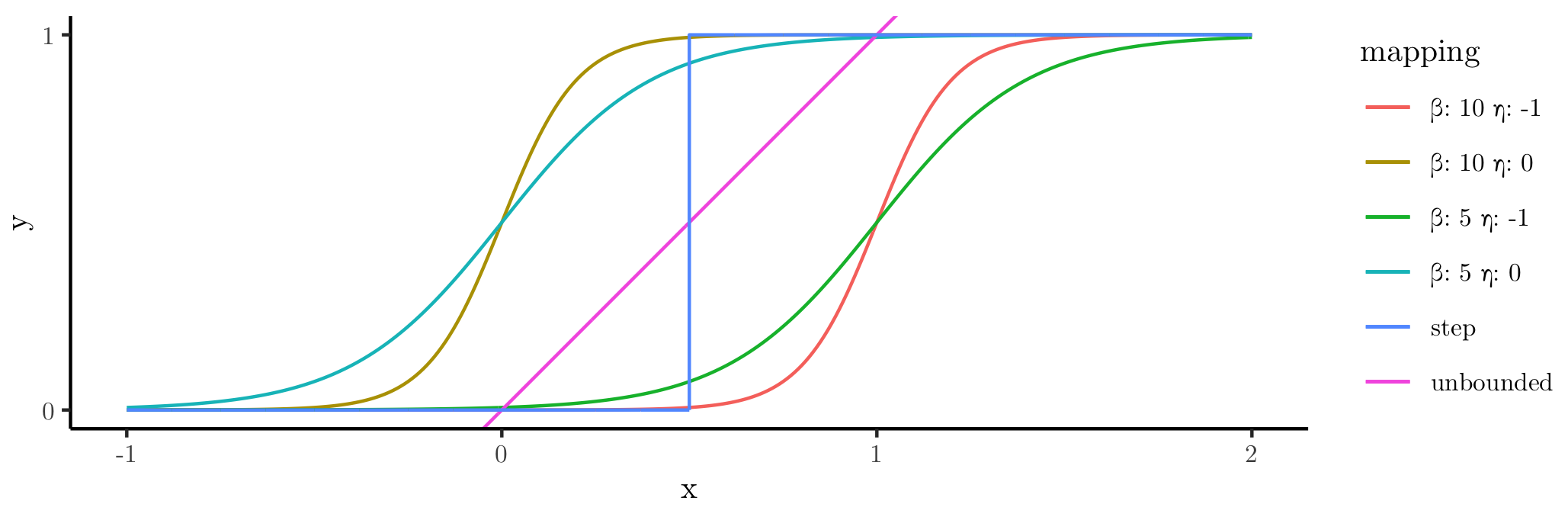}
\caption{Different thresholding regimes: the step function (original $\mathit{F1}$ metric) is not decomposable, the linear function is unbounded ($\mathcal{L}_{\overline{\mathit{F1}}}$) and tends to produce divergent gradients, whereas the sigmoid function ($\mathcal{L}_{\widetilde{\mathit{F1}}}$) is bounded and allows for differentiation due to its smooth curvature, tunable at different parametrizations.}
\label{fig:sigmoid}
\end{figure}

\subsection{Smooth confusion matrix entries}
\label{sec:smoothConf}

We propose a sigmoid-based transformation of the confusion matrix that renders its entries decomposable and fulfills the three desirable properties above:
\begin{equation}
\mbox{}\hspace*{-3mm}\begin{array}{l@{~}l} \widetilde{\mathit{tp}} = \sum \mathbf{S}(\hat{\mathbf{y}}) {\odot} \mathbf{y}  & \widetilde{\mathit{fp}} = \sum \mathbf{S}(\hat{\mathbf{y}}) {\odot} (\mathds{1} - \mathbf{y}) \\[.5em] \widetilde{\mathit{fn}} = \sum (\mathds{1} - \mathbf{S}(\hat{\mathbf{y}})) {\odot} \mathbf{y} & \widetilde{\mathit{tn}} = \sum (\mathds{1} - \mathbf{S}(\hat{\mathbf{y}})) {\odot} (\mathds{1} - \mathbf{y}),
\end{array}\hspace*{-4mm}\mbox{}
\label{eq:smooth}
\end{equation}
with $\mathbf{S}(\cdot)$ the vectorial form of the sigmoid function $S(\cdot)$:
\begin{equation}
S(u; \beta, \eta)=\frac{1}{1+\exp (-\beta (u + \eta))},
\end{equation}

with \(\beta\) and \(\eta\) tunable parameters for slope and offset, respectively. Higher \(\beta\) results in steeper slope at the center of the sigmoid and thus more stringent thresholding. At its extreme, \(\lim_{\beta\to\infty} S(u; \beta, \eta)\) corresponds to the step function used in Eq.~\ref{eq:conf}. Note that negative values of $\beta$ geometrically reflect the sigmoid function across the horizontal line at $0.5$ and thus invert predictions.
These smooth confusion matrix entries allow us to build any related metric (see Table~\ref{tab:confusion-matrix} in Appendix~\ref{sec:evalMetrics}). Furthermore, the surrogate entries are decomposable, bounded, saturated and have a dynamic gradient.

\subsection{Smooth macro F1 scores}
\label{sec:orgc5d29d7}

F1 scores can be calculated on a macro and micro level. Macro-averaging regards all classes as equally important, whereas micro-averaging reflects within-class frequency. \emph{unboundedF1} and \emph{sigmoidF1} below are thought of as macro scores (aggregated over all classes). These scores require a high enough number of representatives in the four confusion matrix quadrants to learn from batch to batch. Ideally, each training epoch would have only one batch, so as to have the most representatives.
Following Eq.~\ref{eq:unbounded}, it is possible to define an \emph{unbounded F1} score:
\begin{equation}\label{eq:unboundedF1}
\mathcal{L}_{\overline{\mathit{F1}}} = 1 - \overline{\mathit{F1}}, \quad \text{where} \quad \overline{\mathit{F1}} = \frac{2 \overline{tp}}{2 \overline{tp}+ \overline{fn}+ \overline{fp}}.
\end{equation}
While this alternative abstracts the thresholding away, which is convenient for fine-tuning purposes, it does not fulfill the desirable properties of a binarization threshold surrogate (see Section~\ref{subsec:unbounded}). \emph{unboundedF1} will be used to benchmark against our proposed \emph{sigmoidF1} loss. Given the definitions of smooth confusion matrix metrics above, we can now write $\mathcal{L}_{\widetilde{\mathit{F1}}}$:
\begin{equation}\label{eq:sigmoidF1}
\mathcal{L}_{\widetilde{\mathit{F1}}} = 1 - \widetilde{\mathit{F1}}, \quad \text{where} \quad \widetilde{\mathit{F1}} = \frac{2\widetilde{\mathit{tp}}}{2 \widetilde{\mathit{tp}}+ \widetilde{\mathit{fn}}+ \widetilde{\mathit{fp}}}.
\end{equation}

\emph{sigmoidF1} is particularly suited for the multilabel setting because it is a proper hard thresholding surrogate as defined in the previous sections and because it contains a significant amount of information about label prediction accuracy: $\widetilde{\mathit{tp}}$, $\widetilde{\mathit{fn}}$ and $\widetilde{\mathit{fp}}$ are indicative of the number of predicted labels in each category of the confusion matrix but also contain a notion of certainty, given that they are rational numbers. The built in sigmoid function ensures that certainty increases along training epochs, as outlined by Property~\ref{prop:sat}. Finally, as the harmonic mean of precision and recall (a property of F1 in general), it weighs in both relevance metrics.

In the next section, we implement Eq.~\ref{eq:sigmoidF1} in PyTorch and TensorFlow as a custom loss as follows:

\begin{lstlisting}[language=Python,mathescape]
# with y the ground truth and z the outcome of the last layer
sig = 1 / (1 + exp(- $\beta$ * (z + $\eta$))) 
tp = sum(sig * y, dim=0)
fp = sum(sig * (1 - y), dim=0)
fn = sum((1 - sig) * y, dim=0)
sigmoid_f1 = 2*tp / (2*tp + fn + fp + 1e-16)
minimize(1 - sigmoid_f1)
\end{lstlisting}

The pseudocode above illustrates the elementwise multiplication of matrices $\mathbf{S}(\mathbf{\hat{y}})$ and $\mathbf{\hat{y}}$ over all examples in the batch and all possible classes.

\section{Experimental Setup}
\label{sec:orgb44ba25}

We test multilabel learning using our proposed \emph{sigmoidF1} loss function on four datasets across different modalities (image and text).
For each modality we take a state-of-the-art model that generates an embedding layer and append a sigmoid activation and different losses. Multilabel deep learning is usually implemented with sigmoid binary cross-entropy directly on the last neural layer (a simplification of the OVA and PAL reductions). We follow this approach for our experiments (e.g., in (large) language models~\citep{bigBird, BERT}). Some baselines include multilabel reformulation choices: only keeping the top-$n$ occurring classes (often 4--10)~\citep[e.g.,][]{textClassificationLeCun, CUNHA}, multiclass classification on each entity within an example (objects in an image, expressions in a text)~\citep[e.g.,][]{COCO, multilabelImage1, multilabelImage2, multilabelImage3}. We refrain from doing so.

\begin{table}[h]
\caption{Descriptive statistics of our experimental datasets.}
\label{table:datasets}
\centering
\begin{tabular}{l l rrr}
\toprule
Dataset & Type & Classes & Average label count & Number of examples \\
\midrule
moviePosters & image & 28 & 2.2 & 37,632\\
arXiv2020 & text & 155 & 1.9 & 26,558\\
Pascal-VOC & image & 20 & 1.6 & 9,963\\
MS-COCO & image & 80 & 2.9 & 122,218\\
\bottomrule
\end{tabular}
\end{table}

\subsection{Datasets}

Table~\ref{table:datasets} lists the datasets we use.
Two of the datasets are multilabel in nature. moviePosters is related to movies \citep{moviePostersData} and arXiv2020 relates to arXiv paper abstracts \citep{arXiv}.
We use the image segmentation datasets Pascal-VOC~\citep{pascalVOC} and MS-COCO~\citep{COCO}, with bounding boxes and one label per box. By attributing all box labels to the image as a whole, it has been used as a reference benchmark for multilabel classification. We refer to Appendix~\ref{app:experimental-setup} for further descriptions of the datasets and references.

\subsection{Learning framework}
\label{subsec:lf}

Our proposed learning framework consists of two parts: a pretrained deep neural network and a classification head (see Figure \ref{fig:architecture}); different loss functions are computed in the classification head.

\textbf{Neural network architecture.} For the moviePoster image dataset, we use a MobileNetV2~\citep{mobileNet} architecture that was pretrained on ImageNet~\citep{imagenet}. This network architecture is typically used for inference on small computing devices (e.g., smartphones). We use a version of MobileNetV2 already stripped off of its original classification head~\citep{mobileNetV2}.
For the three text datasets, we use DistilBERT~\citep{distilBert} as implemented in Hugging Face. This is a particularly efficient instance of the BERT model~\citep{DistilBERTModel}.
For the Pascal-VOC and MS-COCO datasets, we use the recent state-of-the-art resnet TresNet~\citep{TresNet} pretrained on ImageNet~\citep{imagenet} and some of the best practices for Pascal-VOC and MS-COCO collected in a recent benchmark~\citep{ASL}. 
We use TresNet-m-21K; 21K stands for Imagenet21K, the larger ImageNet corpus.
In all cases, we use the final pre-trained layer as an embedding of the input. 
To ensure that the results of different loss functions are comparable, we fix the model weights of the pretrained MobileNetV2, DistilBERT and TresNet and keep the hyperparameter values that were used to be trained from scratch. At training time, we optimize with Adam for all three architectures and use In-Place Activated BatchNorm (Inplace-ABN) for TresNet~\citep{inplaceABN}.

The \textbf{classification head} is a latent representation layer (the final pretrained layer mentioned above) connected with a RELU activation. This layer is linked to a final classification layer with a linear activation. The dimension of the final layer is equal to the number of classes in the dataset. The attached loss function is either BCE (Binary Cross-Entropy), focalLoss~\citep{focalLoss}, ASL~\citep{ASL}, unboundedF1 or sigmoidF1 (ours). When computing the loss at training time, a sigmoid transforms the unbounded last layer to a $[-1,1]$ bounded vector that contrasts positive and negative predictions. These values are then used as inputs to  any of the loss functions above over all classes and the entire batch of examples. In the case of $\mathcal{L}_{\widetilde{\mathit{F1}}}$, this corresponds to a surrogate macro F1. Given the vectorized computation of $\mathcal{L}_{\widetilde{\mathit{F1}}}$ (see Section \ref{sec:smoothConf}), the computational burden is only marginally affected. At inference time, the last layer is used for prediction and is bounded with a sigmoid function. A threshold must then be chosen at evaluation time to compute different metrics. Figure \ref{fig:architecture} depicts this learning framework.

\textbf{Metrics.} In our experiments, we report on microF1, macroF1, Precision, mAP (used in some recent multilabel benchmarks; see Appendix~\ref{sec:evalMetrics}) and (micro-)weightedF1 (where within-class scores are weighted by their representation in the dataset). We focus our discussion around weightedF1 as it is the most comprehensive F1 measure we could find on multilabel problems: it is a micro measure, thus accounts for differences between classes, and has a reweighing argument, thus accounting for class imbalance. Given limited resources we rerun each model on each loss with 5 random seeds. With only 5 runs per loss function, hypothesis testing results would have been particularly sensitive to the choice of distribution.\footnote{We found that, given some unstable results on unboundedF1, even a conservative student t distribution would imply that the $95\%$ confidence interval covers metric values over $100\%$} Instead, we show the distribution of results in Appendix~\ref{app:extended-results}, which show robust statistics (median and interquartile range). Note that cross-validation cannot be performed as Pascal-VOC and MS-COCO have fixed train-validation-test sets.
There is an interaction between our optimization on \emph{sigmoidF1} and our evaluation using (weighted) F1 metrics. We expect higher values on F1-related metrics during evaluation and thus report on alternative metrics too.

\subsection{Hyperparameters and reproducibility}
\label{subsec:hypers}

We implemented all losses in Pytorch and Tensorflow. Batch size is set at a relatively high value of 256 to increase accuracy over traditional losses~\citep{bigBSArxiv}, but also allow heterogeneity in the examples within the batch, thus collecting enough values in each quadrant of the confusion matrix (see Section \ref{sec:orgc5d29d7} for a discussion).
Regarding the \emph{sigmoidF1} hyperparameters $\beta$ and $\eta$, we performed a grid search with the values in the range $[1,30]$ for $\beta$ and $[0, 2]$ for $\eta$.
In our experiments, we evaluate the sensitivity of our method to these hyperparameters (see Figure~\ref{fig:sigmoid} and Appendix~\ref{app:experimental-setup} for optimal values).
We made sure to split the data in the same training, validation and test sets for each loss function. We trained for 60 (Pascal-VOC, MS-COCO) to 100 (arXiv2020, moviePosters) epochs, depending on convergence. Our code, dataset splits and other settings are shared to ensure reproducibility of our results.

\section{Experimental Results}
\label{sec:orgc23a664}

The goal of \emph{sigmoidF1} ($\mathcal{L}_{\widetilde{\mathit{F1}}}$) is to optimize for the F1 score directly at training time in the context of multilabel classification. In this section, we test whether $\mathcal{L}_{\widetilde{\mathit{F1}}}$ can outperform existing loss functions on multiple classification metrics. We present multilabel classification results for $\mathcal{L}_{\widetilde{\mathit{F1}}}$ on four datasets, movie\-Posters, arXiv2020, Pascal-VOC and MS-COCO in Table~\ref{table:overallresults}.

We recall Table~\ref{table:litrev}, in which we highlight that $\mathcal{L}_{\text {BCE}}$ is originally designed for binary classification, $\mathcal{L}_{\text {FL}}$ for imbalanced multiclass, $\mathcal{L}_{\text {ASL}}$ to optimize mAP for multilabel classification. They are computed  over each class at training time, as opposed to per batch for our $\mathcal{L}_{\widetilde{\mathit{F1}}}$ and $\mathcal{L}_{\overline{\mathit{F1}}}$. The latter two explicitly account for label dependencies in the loss function.

In general, Table~\ref{table:overallresults} shows that  $\mathcal{L}_{\widetilde{\mathit{F1}}}$  outperforms other loss functions on three possible formulations of the F1 metric (weightedF1, microF1 and macroF1). We also confirm that the recent ASL loss outperforms other losses on the precision and mAP metrics. $\mathcal{L}_{\widetilde{\mathit{F1}}}$ is designed as an F1 surrogate, it is thus not surprising for it to perform best on F1 metrics and comes at no noticeable additional computational cost (see Appendix~\ref{app:compute-time}). We first analyze the F1 metrics before interpreting the precision and mAP results in more detail.

\header{Measured on the F1 metrics (weightedF1, microF1 and macroF1),} $\mathcal{L}_{\widetilde{\mathit{F1}}}$ and $\mathcal{L}_{\text{BCE}}$ always share the top 2 in performance, oftentimes far ahead of other losses. This highlights that losses inspired by BCE are not yet tailored to optimize for the F1 score in multilabel classification, and also that BCE is a good default choice in general. However, in certain settings, and in particular with our standard datasets Pascal-VOC and MS-COCO, $\mathcal{L}_{\widetilde{\mathit{F1}}}$ can provide clear improvements over the original BCE. macroF1 on the moviePosters dataset is a counter-intuitive exception to that observation: BCE outperforms $\mathcal{L}_{\widetilde{\mathit{F1}}}$ only on the macro measure, although $\mathcal{L}_{\widetilde{\mathit{F1}}}$ is essentially a macro F1 loss function, as it is calculated across all classes and over each entire batch. Similarly focalLoss is dominant on MS-COCO macroF1, but not significantly (see Figure \ref{fig:bxpltCOCO}). There is room for improvement on MS-COCO because we did not finetune the sigmoidF1 hyperparameters  ($\beta$ and $\eta$) and instead reused the Pascal-VOC ones, due to resource constraints.

\header{On precision and mAP,} no top 2 losses emerge. Instead, results are dataset and modality dependent. Surprisingly, the traditional BCE loss outperforms other losses by far in precision on a thoroughly benchmarked dataset like Pascal-VOC. focalLoss delivers best results for MS-COCO on precision, probably because the original paper used MS-COCO as a benchmark to design their loss function~\citep{focalLoss}. Precision performance gains are less clear on the two smaller datasets (arXiv2020 and moviePosters); $\mathcal{L}_{\overline{\mathit{F1}}}$ performs reasonably well.\footnote{$\mathcal{L}_{\overline{\mathit{F1}}}$ was found particularly unstable for Pascal-VOC over 5 different seeds (see the extended results in Appendix~\ref{app:extended-results}). Provided it is unbounded, predictions can diverge towards (positive or negative) infinite values.} Regarding mAP, $\mathcal{L}_{\text {ASL}}$ expectedly outperforms other methods on Pascal-VOC, confirming their own benchmarks~\citep{ASL} and their ability to beat focalLoss and BCE on MS-COCO and PASCAL-VOC. Notably, $\mathcal{L}_{\text {ASL}}$ is also first on mAP on text data. This is the first time that ASL is tested on text data to the best of our knowledge. Overall, these mitigated results for precision and mAP motivate further research in optimizing directly for precision and mAP at training time.

\begin{table}[t]
\caption{Multilabel classification mean performance in percent over 5 random seeds. The F1 metric variants are the focus here (weightedF1, microF1 and macroF1), since we aim to directly optimize for F1 at training time. precision and mAP are displayed for reference, as they are often used in the literature in that context. Metric are formally defined in Appendix~\ref{sec:evalMetrics} and thresholds are indicated there for each dataset. We reused fine-tuned Pascal-VOC sigmoidF1 hyperparameters ($\beta$ and $\eta$) for MS-COCO due to resource constraints.}
\label{table:overallresults}
\centering
\begin{tabular}{ll ccccc}
\toprule
& \textbf{Loss}  & \rotatebox{0}{\textbf{weightedF1}} & \rotatebox{0}{\textbf{microF1}} & \rotatebox{0}{\textbf{macroF1}} & \rotatebox{0}{precision} & \rotatebox{0}{mAP}\\
  \midrule
\multirow{5}{3.7cm}{TresNetm21K [\citeyear{TresNet}]\\ on MS-COCO @0.5 (CNN)} & $\mathcal{L}_{\text {BCE}}$[\citeyear{Fisher}] &      79.02 &   75.81 &   79.55 &     82.52 &  81.21 \\
& $\mathcal{L}_{\text {FL}}$[\citeyear{focalLoss}]    &    81.28 &   79.18 &   \textbf{81.76} &     \textbf{85.73} &  84.88 \\
& $\mathcal{L}_{\text {ASL}}$[\citeyear{ASL}]          &     73.48 &   70.36 &   70.81 &     60.16 &  \textbf{85.59} \\
& $\mathcal{L}_{\overline{\mathit{F1}}}$[ours]  &      79.90 &   77.51 &   79.74 &     81.05 &  78.33 \\
& $\mathcal{L}_{\widetilde{\mathit{F1}}}$[ours]    &     \textbf{81.82} &   \textbf{79.93} &   81.67 &     80.62 &  81.98 \\
\midrule
\multirow{5}{3.7cm}{TresNetm21K [\citeyear{TresNet}]\\ on Pascal-VOC @0.5 (CNN)} & $\mathcal{L}_{\text {BCE}}$[\citeyear{Fisher}] &      87.52 &   85.85 &   87.76 &     \textbf{90.75} &  91.54 \\
& $\mathcal{L}_{\text {FL}}$[\citeyear{focalLoss}]    &      72.54 &   59.24 &   76.82 &     84.70 &  76.19 \\
& $\mathcal{L}_{\text {ASL}}$[\citeyear{ASL}]          &      77.85 &   76.53 &   75.98 &     65.36 &  \textbf{93.11} \\
& $\mathcal{L}_{\overline{\mathit{F1}}}$[ours]  &      77.24 &   74.84 &   75.31 &     75.53 &  79.36 \\
& $\mathcal{L}_{\widetilde{\mathit{F1}}}$[ours]    &      \textbf{88.20} &   \textbf{87.70} &   \textbf{87.87} &     85.36 &  92.36 \\
\midrule
\multirow{5}{3.7cm}{DistilBert [\citeyear{distilBert}]\\ on arXiv2020 @0.05 (NLP)} & $\mathcal{L}_{\text {BCE}}$[\citeyear{Fisher}] &      20.59 &   18.19 &   18.42 &     10.15 &  10.50 \\
& $\mathcal{L}_{\text {FL}}$[\citeyear{focalLoss}]    &      18.85 &   16.59 &   18.01 &     10.10 &  10.43 \\
& $\mathcal{L}_{\text {ASL}}$ [\citeyear{ASL}]         &      19.15 &   16.90 &   18.16 &     \textbf{10.32} &  \textbf{10.53} \\
& $\mathcal{L}_{\overline{\mathit{F1}}}$[ours]  &      15.23 &   13.74 &   14.50 &     10.27 &  10.49 \\
& $\mathcal{L}_{\widetilde{\mathit{F1}}}$[ours]    &      \textbf{20.60} &   \textbf{18.20} &   \textbf{18.43} &     10.15 &  10.50 \\
\midrule
\multirow{5}{3.7cm}{MobileNetV2 [\citeyear{mobileNet}]\\ on moviePosters @0.05 (CNN)} & $\mathcal{L}_{\text {BCE}}$[\citeyear{Fisher}] &      13.79 &    \phantom{1}9.47 &   \textbf{12.94} &      \phantom{1}5.51 &  \phantom{1}5.78 \\
& $\mathcal{L}_{\text {FL}}$[\citeyear{focalLoss}]    &       \phantom{1}0.00 &    \phantom{1}0.00 &      \phantom{1}0.00 &     \phantom{1}0.00 &  \phantom{1}5.80 \\
& $\mathcal{L}_{\text {ASL}}$[\citeyear{ASL}]          &       \phantom{1}0.00 &    \phantom{1}0.00 &      \phantom{1}0.00 &      \phantom{1}0.00 &  \phantom{1}5.80 \\
& $\mathcal{L}_{\overline{\mathit{F1}}}$[ours]  &      13.97 &    \phantom{1}9.84 &   10.11 &      \phantom{1}\textbf{5.59} &  \phantom{1}\textbf{5.90} \\
& $\mathcal{L}_{\widetilde{\mathit{F1}}}$[ours]    &      \textbf{14.81} &   \textbf{10.33} &   10.57 &      \phantom{1}5.58 &  \phantom{1}5.81 \\
\bottomrule
\end{tabular}
\end{table}

\header{A note on thresholding and zero values.} For the bigger and more standard datasets Pascal-VOC and MS-COCO,\footnote{The classes in Pascal-VOC and MS-COCO are a lot more concrete (e.g., car, person, bicycle) and are directly related to the original classes of ImageNet on which the TresNet and MobileNetV2 were trained, as opposed to movie genres for moviePosters or arXiv paper scientific domain.} our neutral metric threshold of $0.5$ provides results in line with the literature. With our own fine-tuning regime on a smaller model (see Section \ref{subsec:lf}), our mAP scores are $1$--$2\%$ away from the current state of the art~\citep{ASL}. On smaller datasets like arXiv2020, moviePosters and others (see Appendix~\ref{app:addtional-experiments}), the sigmoid activation per class at inference time are closer to zero. To a certain extent, this can be interpreted as the model having less confidence in its predictions~\citep{predsConf}. As a result, a neutral $0.5$ threshold resulted in zero values on almost all losses and metrics for small datasets. Given the range of values in these predictions, $0.05$ seems like the next best neutral threshold. We refrain from further finetuning the threshold for each dataset, loss and metric.\footnote{While optimizing the threshold at inference time is an interesting research topic, we refrain from doing so here, so as to disentangle the loss function benchmarking from the thresholding regime benchmarking.} As a result of the absence of finetuning, moviePosters display zero values for $\mathcal{L}_{\text {FL}}$ and $\mathcal{L}_{\text {ASL}}$ on most metrics. This can be explained by the higher average label count for moviePosters. This is in opposition to the propensity of $\mathcal{L}_{\text {FL}}$ and $\mathcal{L}_{\text {ASL}}$ to deal with sparser label representation.

The analysis above highlights that sigmoidF1 can indeed optimize for F1 metrics (weightedF1, microF1 and macroF1) reliably and consistently, over six datasets in total (see Appendix~\ref{app:extended-results}). Given the more mitigated results for precision and mAP, it seems relevant to further explore opportunities of metrics-as-losses. Finally, BCE, which was designed with binary classification in mind, is a good first approximation.

\begin{figure*}[t!]
\centering
\includegraphics[width=.85\linewidth]{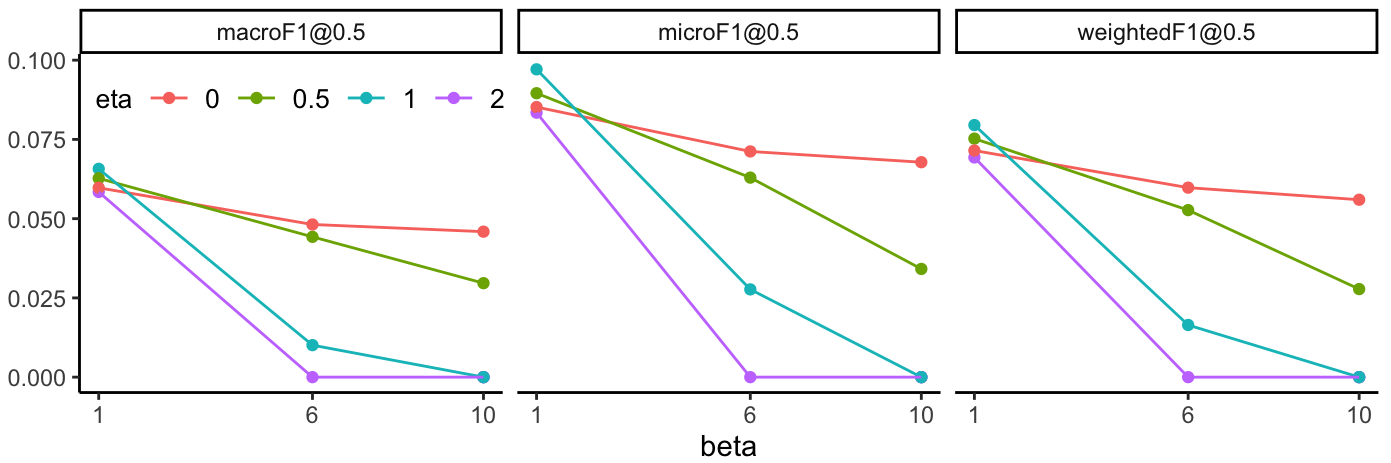}
\vspace{.1\baselineskip}
\caption{DistilBERT (NLP) on arXiv2020 – different weightedF1 scores at a 0.5 threshold for different values of $\eta$ and $\beta$ in a sampling region similar to Figure~\ref{fig:sigmoid}.}
\label{fig:betaEta}
\end{figure*}

\header{Sensitivity analysis.}
In Figure~\ref{fig:betaEta}, we show the sensitivity of \emph{sigmoidF1} to different para\-metrizations of $\eta$ and $\beta$. Within the chosen values (see Section~\ref{subsec:hypers}), we chose to display a parameter space similar to the one illustrated in Figure~\ref{fig:sigmoid}. Moving the sigmoid to the left allows the learning algorithm to tend to a (local) optimum.
In general and across datasets, when sampling for $\eta$, we noticed how the optimum tended towards positive values. Offsetting the sigmoid curve to the left has the effect of pushing more candidate predictions to the rank of positive instance (or at least close to 1). We also note how $\beta$ (which cannot be negative or otherwise the sigmoid function would flip around the horizontal axis) is at best close to a value close to 0 on this dataset (we show discrete values here for display purposes). The sigmoid is thus relatively smooth, which involves dynamic gradients over different batches. The idea is similar to a high learning rate. In our experiments, this rarely gave rise to divergent behavior in the loss function (learning curve). We learn that it is necessary to tune hyperparameters for each dataset, as it is for $\mathcal{L}_{\text {FL}}$, $\mathcal{L}_{\text {ASL}}$ and others in Table~\ref{table:litrev}.

The results in this section show that, in general, multilabel classification results measured on F1 metrics can be improved using sigmoidF1 – independently of the dataset, its modality or the neural network architecture.

\section{Discussion}
\label{sec:discussion}

In multilabel classification, and more generally in the context of deep neural networks, losses are formulated to be decomposable for gradient descent. At inference time, however, end-users tend to look for clear-cut actionable decisions from the model (e.g., to automize the arXiv keywords selection, one needs to obtain a clear-cut set of keywords given each abstract). This is probably why most evaluation metrics in the multilabel literature, with the notable exception of mAP, are also reliant on clear-cut counts (e.g. $tp$, $fn$, $fp$, $tn$). Although models are benchmarked on these values, we found little discussions on how to retrieve clear-cut counts from final softmax / sigmoid activations bounded by $[0,1]$. Among our benchmarked losses, the authors of  FocalLoss~\citep{focalLoss} use a global $0.5$ threshold at inference-time. The authors of ASL~\citep{tencent} do not mention thresholding in the paper but a GitHub issue hints at the fact that they used $0.8$ as a global threshold for MS-COCO.\footnote{See https://github.com/Alibaba-MIIL/ASL/issues/8 and also insightful learning tricks at https://github.com/Alibaba-MIIL/ASL/issues/30} We feel that defining clear-cut counts deserves more attention.

Such clear-cut counts are usually achieved via a \emph{decision threshold}. \citet{deepF} distinguish between \emph{utility maximization} (at inference-time) and \emph{decision-theoretic} (at training and at inference time) approaches.

\header{Utility maximization methods.}
At inference time, a threshold can be set globally for all examples to optimize on the training data, before using it on the test data~\citep{threshForF1, deepF}. This threshold can optimize for specificity, sensitivity~\citep{decisionThreshold} or directly for F1~\citep{deepF}. Alternatively, different thresholds can be set per-class~\citep{moviePosters}.

\header{Decision-theoretic methods.}
Decision-theoretic methods operate both at training and at inference time. They stem from shallow learning fields and have multiple steps: 
\begin{enumerate*}[label=(\roman*)]
\item \emph{encoding} the original item-label matrix to submatrices that each can be 
\item fit a traditional loss function (cross-entropy variations), before 
\item \emph{decoding} the submatrices back to the original item-label matrix format via an inference-phase optimization solver. 
\end{enumerate*}
This methodology can be found across the shallow learning fields of SVMs~\citep{shallowF1}, logistic regession~\citep{shallowF2, shallowF5}, multinomial regression~\citep{shallowF3}, and Bayesian networks~\citep{shallowF4}. These methods were implemented in a deep learning setting, where they had more success than \emph{utility maximization} or fixed thresholding methods~\citep{deepF}. Decision-theoretic methods have at least 3 moving parts mentioned above and are thus complicated to benchmark against each other, let alone against inference-time thresholding or fix thresholding.

As hinted before with ASL and focalLoss, modern deep learning models tend to not tune the decision threshold, either with \emph{utility maximization} or \emph{decision-theoretic} methods. We propose to take a first step in this direction in the following.

For the sake of this discussion, we focus on the simplest \emph{utility maximization} (inference-time) thresholding implementation. \emph{Threshold Averaging}~\citep{deepF} is a method that uses the training set to tune a global threshold, before applying it to the test set. Take $\hat{\mathbf{y}}_i$, a set of label predictions for one example $i$. We select each $\hat{y}_{ij}$ as a possible thresholding candidate to binarise the vector $\hat{\mathbf{y}}_i$. We then calculate instance-wise F1 scores over $\hat{\mathbf{y}}_i$. The value $\hat{y}_{ij}$ that results in the highest F1 score for an instance is chosen as the instance's threshold. This process is repeated for each instance $i$ in the training data. The average threshold over all instances in the training data is then chosen as the final global threshold for the test data.

In Table \ref{table:avgThresh}, we show results of \emph{Threshold Averaging}~\citep{deepF} on the arXiv dataset. It is notable here that ASL's mean results always outperform other losses. This time around, however, almost all boxplots IQRs intersect, thus results are very inconclusive (see Figure~\ref{fig:bxpltArxivAvgThresh}). We thus refrain from bold numbers like in Table~\ref{table:overallresults}. Most importantly, metrics are consistently below results from the original neutral fixed 0.05 threshold in Table~\ref{table:overallresults}. This is consistent with some of the results in \citep{deepF}, showing that simple thresholding methods based on \emph{utility maximization} are not sufficient to consistently beat fixed thresholds or \emph{decision-theoretic} methods.

Inference-time decisions can completely change the outcome of a prediction set, of its resulting evaluation metrics, and, thus, even of the winning model. We hope that thresholding will be more broadly discussed in the future or at least for the thresholding method to be openly stated in research papers; we chose fixed neutral thresholds, to focus on the benchmarking of losses at training-time.

Together, utility maximization (inference-time) thresholding methods and decision-theoretic methods (at training and at inference time) form an under-explored research domain, with several open questions:
\begin{enumerate*}[label=(\roman*)]
\item Which data split should be used for thresholding? With the entire training dataset~\citep{deepF}, there is a risk of overfitting the threshold. Maybe it is worth introducing a holdout set that is only used for threshold tuning.
\item Should we threshold globally for interpretability or have a per-class or even per-instance threshold?
\item Are \emph{decision theoretic} (a.k.a. at training and at inference time) approaches also not prone to overfitting and are they efficient on large neural networks for large datasets?
\item Can other losses than the classical cross-entropy loss be used to train \emph{decision theoretic} models?
\end{enumerate*}

\begin{table}[t]
  \caption{Multilabel classification mean performance in percent over 5 random seeds. Global thresholds were found using a threshold-moving technique. Results are systematically lower than with a fixed threshold (see third row of Table~\ref{table:overallresults}). Metric are formally defined in Appendix~\ref{sec:evalMetrics}.}
\label{table:avgThresh}
\centering

\begin{tabular}{ll ccccc}
\toprule
& \textbf{Loss}  & \rotatebox{0}{\textbf{weightedF1}} & \rotatebox{0}{\textbf{microF1}} & \rotatebox{0}{\textbf{macroF1}} & \rotatebox{0}{precision} & \rotatebox{0}{mAP}\\
  \midrule

\multirow{5}{3.7cm}{DistilBert [\citeyear{distilBert}]\\ on arXiv2020 (NLP) – Threshold moving} & 
$\mathcal{L}_{\text {BCE}}$[\citeyear{Fisher}] &      15.89 &   13.98 &   15.73 &     10.11 &  10.35  \\
& $\mathcal{L}_{\text {FL}}$[\citeyear{focalLoss}]    &      16.40 &   14.14 &   17.22 &      9.83 &  10.42  \\
& $\mathcal{L}_{\text {ASL}}$ [\citeyear{ASL}]         &      17.49 &   14.86 &   17.77 &     10.33 &  10.51  \\
& $\mathcal{L}_{\overline{\mathit{F1}}}$[ours]  &      16.52 &   14.27 &   16.70 &      9.98 &  10.43  \\
& $\mathcal{L}_{\widetilde{\mathit{F1}}}$[ours]    &      15.11 &   13.19 &   15.20 &     10.05 &  10.41 \\  

\bottomrule
\end{tabular}
\end{table}

\section{Conclusions}
\label{sec:orged3d8a1}

To solve multilabel learning tasks, existing optimization frameworks are typically based on variations of the cross-entropy loss. Instead – inspired by the binary classification literature (see most recently \citep{multiclassOptDL2} and their F1 surrogate loss functions on 3-layer neural networks) – we propose the \emph{sigmoidF1} loss, as part of a general loss framework for confusion matrix metrics. \emph{sigmoidF1} loss can achieve significantly better results for most metrics on four diverse datasets and outperforms other losses on the weightedF1 metric. We thereby provide evidence that \emph{sigmoidF1} is robust to modality, model architecture and dataset size, when optimizing for F1 metrics.
Generally, our smooth formulation of confusion matrix metrics allows us to optimize directly for these metrics that are usually reserved for the evaluation phase. The proposed \emph{unboundedF1} counterpart does not require hyperparameter tuning and delivered better results than existing multiclass losses on most metrics; it can act as a mathematically less robust approximation of \emph{sigmoidF1}.

In future work and within the generic multilabel setting, a first incremental step could be to train on a bigger dataset like MS-COCO~\citep{COCO} (if provided with more resources) and use more robust transfer learning/finetuning procedures, for example with dynamic weight freezing for finetuning~\citep{ULMFit}. Alternatively, we could train a CNN or a BERT model for multilabel tasks with our smooth losses from scratch (cf., \citep{tencent} and \citep{focalLoss}). If training from scratch, this can be combined with representation learning \citep{unsupervisedImage,highResRepresentation} or self-supervised learning, in order to model abstract relationships.

Next, we could validate if F1 or another confusion-matrix-metric-as-a-loss can tackle other multilabel settings, such as hierarchical multilabel classification \citep{HARAM}, active learning \citep{activeLearningMultiLabel}, multi-instance learning~\citep[e.g.,][]{multiInstance,multiInstanceMultiLabel}, holistic label learning (see dataset \emph{Large Scale Holistic Video Understanding}~\citep{holisticVideoData}), or
extreme multilabel prediction \citep{extremeMultilabelText, extremeSIGIR, extremeDiSMEC, extremePPDsparse, extremeParabel} (with missing labels~\citep{exteremeMissing, exteremeMissingApplications}), where the number of classes ranges in the tens of thousands. Beyond the multilabel setting, \emph{sigmoidF1} could be tested on any model that uses F1 score as an evaluation metric such as AC-SUM-GAN \citep{AC-SUM-GAN}.

One limitation of \emph{sigmoidF1} is that it is computed at a macro level over the whole batch and ignores (micro) per class F1 scores. Given our limited GPU memory, we could not load enough examples in each batch to represent each confusion matrix quadrant of each class reliably. If such a route is followed, we could eventually finetune or learn $\beta_c$ and $\eta_c$ – the parameters of the sigmoid function – per class $c$.

We believe that smooth metric surrogates should inform future research on multilabel classification tasks. There is evidence of a growing interest in the literature~\citep{softF1, sigmoid, recallLoss} for metrics as losses and the objective of this paper is to further highlight their relevance, across modalities, architectures and dataset sizes. Based on the results presented in this paper, we consider metrics-as-losses (e.g., Jaccard, confusion matrix metrics, ranking metrics) as the next step in the evolution of multilabel classification algorithms.

\subsubsection*{Acknowledgments}
We thank our reviewers and the action editor for their valuable comments and suggestions, especially for pointing out to the thresholding literature, which led us to add a discussion section. We thank Nanne van Noord for his valuable suggestions on the paper structure and Maurits Bleeker for reviewing the text.

This research was (partially) funded by Bertelsmann SE \& Co. KGaA; by the Hybrid Intelligence Center, a 10-year program funded by the Dutch Ministry of Education, Culture and Science through the Netherlands Organisation for Scientific Research, \url{https://hybrid-intelligence-centre.nl}.

All content represents the opinion of the authors, which is not necessarily shared or endorsed by their respective employers and/or sponsors.

\subsubsection*{Author Contributions}
All authors participated in the ideation process and discussions. Gabriel Bénédict was responsible for first drafts, the code and visualizations (except Figure 1, which is mostly attributable to Daan Odijk). Hendrik Vincent Koops, Daan Odijk and Maarten de Rijke revised initial drafts, proposed changes and feedback in content and form.

\bibliography{references.bib}

\begin{thebibliography}{115}
\providecommand{\natexlab}[1]{#1}
\providecommand{\url}[1]{\texttt{#1}}
\expandafter\ifx\csname urlstyle\endcsname\relax
  \providecommand{\doi}[1]{doi: #1}\else
  \providecommand{\doi}{doi: \begingroup \urlstyle{rm}\Url}\fi

\bibitem[Abadi et~al.(2015)Abadi, Agarwal, Barham, Brevdo, Chen, Citro,
  Corrado, Davis, Dean, Devin, Ghemawat, Goodfellow, Harp, Irving, Isard, Jia,
  Jozefowicz, Kaiser, Kudlur, Levenberg, Man\'{e}, Monga, Moore, Murray, Olah,
  Schuster, Shlens, Steiner, Sutskever, Talwar, Tucker, Vanhoucke, Vasudevan,
  Vi\'{e}gas, Vinyals, Warden, Wattenberg, Wicke, Yu, and Zheng]{tensorflow}
Mart\'{\i}n Abadi, Ashish Agarwal, Paul Barham, Eugene Brevdo, Zhifeng Chen,
  Craig Citro, Greg~S. Corrado, Andy Davis, Jeffrey Dean, Matthieu Devin,
  Sanjay Ghemawat, Ian Goodfellow, Andrew Harp, Geoffrey Irving, Michael Isard,
  Yangqing Jia, Rafal Jozefowicz, Lukasz Kaiser, Manjunath Kudlur, Josh
  Levenberg, Dandelion Man\'{e}, Rajat Monga, Sherry Moore, Derek Murray, Chris
  Olah, Mike Schuster, Jonathon Shlens, Benoit Steiner, Ilya Sutskever, Kunal
  Talwar, Paul Tucker, Vincent Vanhoucke, Vijay Vasudevan, Fernanda Vi\'{e}gas,
  Oriol Vinyals, Pete Warden, Martin Wattenberg, Martin Wicke, Yuan Yu, and
  Xiaoqiang Zheng.
\newblock {TensorFlow}: Large-scale machine learning on heterogeneous systems,
  2015.
\newblock Software available from tensorflow.org.

\bibitem[Agrawal et~al.(2013)Agrawal, Gupta, Prabhu, and
  Varma]{millionsOfLabels}
Rahul Agrawal, Archit Gupta, Yashoteja Prabhu, and Manik Varma.
\newblock Multi-label learning with millions of labels: Recommending advertiser
  bid phrases for web pages.
\newblock In \emph{Proceedings of the 22nd International Conference on World
  Wide Web}, WWW '13, pp.\  13--24, New York, NY, USA, 2013. Association for
  Computing Machinery.

\bibitem[Amoualian et~al.(2020)Amoualian, Goswami, Ach, Das, and
  Montalvo]{Amoualian2020SIGIR2E}
Hesam Amoualian, Parantapa Goswami, Laurent Ach, Pradipto Das, and Pablo
  Montalvo.
\newblock Sigir 2020 e-commerce workshop data challenge overview.
\newblock In \emph{Proceedings Proceedings of ACM SIGIR Workshop on eCommerce
  (SIGIR eCom'20). ACM}. ACM, 2020.

\bibitem[Apostolidis et~al.(2020)Apostolidis, Adamantidou, Metsai, Mezaris, and
  Patras]{AC-SUM-GAN}
Evlampios Apostolidis, Eleni Adamantidou, Alexandros~I. Metsai, Vasileios
  Mezaris, and Ioannis Patras.
\newblock {AC-SUM-GAN}: Connecting actor-critic and generative adversarial
  networks for unsupervised video summarization,.
\newblock \emph{IEEE Transactions on Circuits and Systems for Video
  Technology}, 2020.

\bibitem[Babbar \& Sch\"{o}lkopf(2017)Babbar and Sch\"{o}lkopf]{extremeDiSMEC}
Rohit Babbar and Bernhard Sch\"{o}lkopf.
\newblock Dismec: Distributed sparse machines for extreme multi-label
  classification.
\newblock In \emph{Proceedings of the Tenth ACM International Conference on Web
  Search and Data Mining}, WSDM '17, pp.\  721--729, New York, NY, USA, 2017.
  Association for Computing Machinery.

\bibitem[Baruch et~al.(2020)Baruch, Ridnik, Zamir, Noy, Friedman, Protter, and
  Zelnik{-}Manor]{ASL}
Emanuel~Ben Baruch, Tal Ridnik, Nadav Zamir, Asaf Noy, Itamar Friedman, Matan
  Protter, and Lihi Zelnik{-}Manor.
\newblock Asymmetric loss for multi-label classification.
\newblock \emph{CoRR}, abs/2009.14119, 2020.

\bibitem[Behera et~al.(2019)Behera, {Kumaravelan}, and {Kumar
  B.}]{weightedMetrics}
Bichitrananda Behera, G.~{Kumaravelan}, and P.~{Kumar B.}
\newblock Performance evaluation of deep learning algorithms in biomedical
  document classification.
\newblock In \emph{2019 11th International Conference on Advanced Computing
  (ICoAC)}, pp.\  220--224, 2019.

\bibitem[Benites \& Sapozhnikova(2015)Benites and Sapozhnikova]{HARAM}
Fernando Benites and Elena Sapozhnikova.
\newblock {HARAM}: A hierarchical {ARAM} neural network for large-scale text
  classification.
\newblock In \emph{2015 IEEE International Conference on Data Mining Workshop
  (ICDMW)}, pp.\  847--854, 2015.

\bibitem[Bhatia et~al.(2015)Bhatia, Jain, Kar, Varma, and
  Jain]{extremeMultilabelEmbeddings}
Kush Bhatia, Himanshu Jain, Purushottam Kar, Manik Varma, and Prateek Jain.
\newblock Sparse local embeddings for extreme multi-label classification.
\newblock In C.~Cortes, N.~Lawrence, D.~Lee, M.~Sugiyama, and R.~Garnett
  (eds.), \emph{Advances in Neural Information Processing Systems}, volume~28.
  Curran Associates, Inc., 2015.

\bibitem[Bishop(2007)]{Bishop}
Christopher~M. Bishop.
\newblock \emph{Pattern Recognition and Machine Learning}.
\newblock Information science and statistics. Springer, 5 edition, 2007.

\bibitem[Blosseville et~al.(1992)Blosseville, H\'{e}brail, Monteil, and
  P\'{e}not]{documentClassification}
M.~J. Blosseville, Georges H\'{e}brail, Marie-Ga{\"e}lle Monteil, and
  N.~P\'{e}not.
\newblock Automatic document classification: Natural language processing,
  statistical analysis, and expert system techniques used together.
\newblock In \emph{Proceedings of the 15th Annual International ACM SIGIR
  Conference on Research and Development in Information Retrieval}, SIGIR '92,
  pp.\  51--58, New York, NY, USA, 1992. Association for Computing Machinery.

\bibitem[Boutell et~al.(2004)Boutell, Luo, Shen, and Brown]{labelPowerset}
Matthew~R. Boutell, Jiebo Luo, Xipeng Shen, and Christopher~M. Brown.
\newblock Learning multi-label scene classification.
\newblock \emph{Pattern Recognition}, 37\penalty0 (9):\penalty0 1757--1771,
  2004.

\bibitem[Brinker et~al.(2006)Brinker, F\"{u}rnkranz, and H\"{u}llermeier]{OVA1}
Klaus Brinker, Johannes F\"{u}rnkranz, and Eyke H\"{u}llermeier.
\newblock A unified model for multilabel classification and ranking.
\newblock In \emph{Proceedings of the 2006 Conference on ECAI 2006: 17th
  European Conference on Artificial Intelligence August 29 -- September 1,
  2006, Riva Del Garda, Italy}, pp.\  489--493, NLD, 2006. IOS Press.

\bibitem[Bruno et~al.(2013)Bruno, Sasa, and Donko]{textCategorization}
Trstenjak Bruno, Mikac Sasa, and Dzenana Donko.
\newblock {KNN} with {TF-IDF} based framework for text categorization.
\newblock \emph{Procedia Engineering}, 69:\penalty0 1356--1364, 11 2013.

\bibitem[Chang et~al.(2019)Chang, Yu, Zhong, Yang, and Dhillon]{softF1}
Wei-Cheng Chang, Hsiang-Fu Yu, Kai Zhong, Yiming Yang, and Inderjit~S. Dhillon.
\newblock The unknown benefits of using a soft-f1 loss in classification
  systems.
\newblock \emph{Towards Data Science}, Dec 2019.
\newblock URL
  \url{https://towardsdatascience.com/the-unknown-benefits-of-using-a-soft-f1-loss-in-classification-systems-753902c0105d}.

\bibitem[Chang et~al.(2020)Chang, Yu, Zhong, Yang, and
  Dhillon]{extremeMultilabelText}
Wei-Cheng Chang, Hsiang-Fu Yu, Kai Zhong, Yiming Yang, and Inderjit~S. Dhillon.
\newblock Taming pretrained transformers for extreme multi-label text
  classification.
\newblock \emph{Proceedings of the 26th ACM SIGKDD International Conference on
  Knowledge Discovery \& Data Mining}, Jul 2020.

\bibitem[Chen et~al.(2006)Chen, Tsai, Moon, Ahn, Young, and
  Chen]{decisionThreshold}
James~J. Chen, Chen-An Tsai, Hojin Moon, Hongshik Ahn, John~J. Young, and
  Chun-Houh Chen.
\newblock Decision threshold adjustment in class prediction.
\newblock \emph{SAR and QSAR in Environmental Research}, 17\penalty0
  (3):\penalty0 337--352, Jun 2006.

\bibitem[Chen et~al.(2009)Chen, Liu, Lan, Ma, and Li]{rankingNDCG}
Wei Chen, Tie-yan Liu, Yanyan Lan, Zhi-ming Ma, and Hang Li.
\newblock Ranking measures and loss functions in learning to rank.
\newblock In Y.~Bengio, D.~Schuurmans, J.~Lafferty, C.~Williams, and A.~Culotta
  (eds.), \emph{Advances in Neural Information Processing Systems}, volume~22.
  Curran Associates, Inc., 2009.
\newblock URL
  \url{https://proceedings.neurips.cc/paper/2009/file/2f55707d4193dc27118a0f19a1985716-Paper.pdf}.

\bibitem[Chu \& Guo(2017)Chu and Guo]{moviePosters}
Wei-Ta Chu and Hung-Jui Guo.
\newblock Movie genre classification based on poster images with deep neural
  networks.
\newblock \emph{Proceedings of the Workshop on Multimodal Understanding of
  Social, Affective and Subjective Attributes}, Oct 2017.

\bibitem[Clare \& King(2001)Clare and King]{ML-DT}
Amanda Clare and Ross~D. King.
\newblock Knowledge discovery in multi-label phenotype data.
\newblock \emph{Lecture Notes in Computer Science}, pp.\  42--53, 2001.

\bibitem[Cornell-University(2021)]{arXiv}
Cornell-University.
\newblock {arXiv} dataset and metadata of {1.7M+} scholarly papers across
  {STEM}.
\newblock \url{kaggle.com/Cornell-University/arxiv}, 2021.

\bibitem[Cortes \& Vapnik(1995)Cortes and Vapnik]{softMargin}
Corinna Cortes and Vladimir Vapnik.
\newblock Support-vector networks.
\newblock In \emph{Machine Learning}, pp.\  273--297, 1995.

\bibitem[Cunha et~al.(2021)Cunha, Mangaravite, Gomes, Canuto, Resende,
  Nascimento, Viegas, Fran{\c c}a, Martins, Almeida, Rosa, Rocha, and Gon{\c
  c}alves]{CUNHA}
Washington Cunha, V{\'\i}tor Mangaravite, Christian Gomes, S{\'e}rgio Canuto,
  Elaine Resende, Cecilia Nascimento, Felipe Viegas, Celso Fran{\c c}a,
  Wellington~Santos Martins, Jussara~M. Almeida, Thierson Rosa, Leonardo Rocha,
  and Marcos~Andr{\'e} Gon{\c c}alves.
\newblock On the cost-effectiveness of neural and non-neural approaches and
  representations for text classification: A comprehensive comparative study.
\newblock \emph{Information Processing \& Management}, 58\penalty0
  (3):\penalty0 102481, 2021.

\bibitem[Decubber et~al.(2018)Decubber, Mortier, Dembczynski, and
  Waegeman]{deepF}
Stijn Decubber, Thomas Mortier, Krzysztof Dembczynski, and Willem Waegeman.
\newblock Deep f-measure maximization in multi-label classification: A
  comparative study.
\newblock In \emph{ECML/PKDD}, pp.\  290--305, 2018.
\newblock URL \url{https://doi.org/10.1007/978-3-030-10925-7_18}.

\bibitem[Dembczy\'{n}ski et~al.(2010)Dembczy\'{n}ski, Cheng, and
  H\"{u}llermeier]{OVA2}
Krzysztof Dembczy\'{n}ski, Weiwei Cheng, and Eyke H\"{u}llermeier.
\newblock Bayes optimal multilabel classification via probabilistic classifier
  chains.
\newblock In \emph{Proceedings of the 27th International Conference on
  International Conference on Machine Learning}, ICML'10, pp.\  279--286,
  Madison, WI, USA, 2010. Omnipress.

\bibitem[Dembczynski et~al.(2011)Dembczynski, Waegeman, Cheng, and
  H\"{u}llermeier]{shallowF2}
Krzysztof Dembczynski, Willem Waegeman, Weiwei Cheng, and Eyke H\"{u}llermeier.
\newblock An exact algorithm for f-measure maximization.
\newblock In J.~Shawe-Taylor, R.~Zemel, P.~Bartlett, F.~Pereira, and K.Q.
  Weinberger (eds.), \emph{Advances in Neural Information Processing Systems},
  volume~24. Curran Associates, Inc., 2011.
\newblock URL
  \url{https://proceedings.neurips.cc/paper/2011/file/71ad16ad2c4d81f348082ff6c4b20768-Paper.pdf}.

\bibitem[Dembczynski et~al.(2013)Dembczynski, Jachnik, Kotlowski, Waegeman, and
  Huellermeier]{shallowF3}
Krzysztof Dembczynski, Arkadiusz Jachnik, Wojciech Kotlowski, Willem Waegeman,
  and Eyke Huellermeier.
\newblock Optimizing the f-measure in multi-label classification: Plug-in rule
  approach versus structured loss minimization.
\newblock In Sanjoy Dasgupta and David McAllester (eds.), \emph{Proceedings of
  the 30th International Conference on Machine Learning}, volume~28 of
  \emph{Proceedings of Machine Learning Research}, pp.\  1130--1138, Atlanta,
  Georgia, USA, 17--19 Jun 2013. PMLR.
\newblock URL \url{https://proceedings.mlr.press/v28/dembczynski13.html}.

\bibitem[Deng et~al.(2009)Deng, Dong, Socher, Li, Li, and Fei-Fei]{imagenet}
Jia Deng, Wei Dong, Richard Socher, Li-Jia Li, Kai Li, and Li~Fei-Fei.
\newblock {ImageNet}: A large-scale hierarchical image database.
\newblock In \emph{2009 IEEE Conference on Computer Vision and Pattern
  Recognition}, pp.\  248--255, 2009.

\bibitem[Devlin et~al.(2019)Devlin, Chang, Lee, and Toutanova]{BERT}
Jacob Devlin, Ming-Wei Chang, Kenton Lee, and Kristina Toutanova.
\newblock {BERT}: Pre-training of deep bidirectional transformers for language
  understanding.
\newblock In \emph{Proceedings of the 2019 Conference of the North {A}merican
  Chapter of the Association for Computational Linguistics: Human Language
  Technologies, Volume 1 (Long and Short Papers)}, pp.\  4171--4186,
  Minneapolis, Minnesota, June 2019. Association for Computational Linguistics.

\bibitem[Diba et~al.(2019)Diba, Fayyaz, Sharma, Paluri, Gall, Stiefelhagen, and
  Gool]{holisticVideoData}
Ali Diba, Mohsen Fayyaz, Vivek Sharma, Manohar Paluri, Jurgen Gall, Rainer
  Stiefelhagen, and Luc~Van Gool.
\newblock {Large Scale Holistic Video Understanding}.
\newblock \emph{arXiv preprint arXiv:1904.11451v3}, 2019.

\bibitem[Du et~al.(2019)Du, Chen, Peng, Xiang, Tao, and Lu]{multitaskLabel}
Jingcheng Du, Qingyu Chen, Yifan Peng, Yang Xiang, Cui Tao, and Zhiyong Lu.
\newblock Ml-net: Multi-label classification of biomedical texts with deep
  neural networks.
\newblock \emph{Journal of the American Medical Informatics Association},
  26\penalty0 (11):\penalty0 1279--1285, Jun 2019.

\bibitem[Eban et~al.(2017)Eban, Schain, Mackey, Gordon, Rifkin, and
  Elidan]{optimizableLosses}
Elad Eban, Mariano Schain, Alan Mackey, Ariel Gordon, Ryan Rifkin, and Gal
  Elidan.
\newblock {Scalable Learning of Non-Decomposable Objectives}.
\newblock In Aarti Singh and Jerry Zhu (eds.), \emph{Proceedings of the 20th
  International Conference on Artificial Intelligence and Statistics},
  volume~54 of \emph{Proceedings of Machine Learning Research}, pp.\  832--840.
  PMLR, 20--22 Apr 2017.

\bibitem[Elisseeff \& Weston(2001)Elisseeff and Weston]{multilabelSVM}
Andr\'{e} Elisseeff and Jason Weston.
\newblock A kernel method for multi-labelled classification.
\newblock In \emph{Proceedings of the 14th International Conference on Neural
  Information Processing Systems: Natural and Synthetic}, NIPS'01, pp.\
  681--687, Cambridge, MA, USA, 2001. MIT Press.

\bibitem[Everingham et~al.(2007)Everingham, Gool, Williams, Winn, and
  Zisserman]{pascalVOC}
Mark Everingham, Luc~Van Gool, Christopher~KI Williams, John Winn, and Andrew
  Zisserman.
\newblock The {PASCAL} visual object classes challenge 2007 ({VOC2007})
  results.
\newblock
  \url{http://www.pascal-network.org/challenges/VOC/voc2007/index.html}, 2007.

\bibitem[Fisher(1912)]{Fisher}
Ronald~A. Fisher.
\newblock On an absolute criterion for fitting frequency curves.
\newblock \emph{Messenger of Mathematics}, 41:\penalty0 155--160, 1912.

\bibitem[Gai et~al.(2019)Gai, Zhang, and Cho]{multiclassOptDL2}
Yu~Gai, Zheng Zhang, and Kyunghyun Cho.
\newblock Gradient-based learning for f-measure and other performance metrics,
  2019.
\newblock URL \url{https://openreview.net/forum?id=H1zxjsCqKQ}.

\bibitem[Gasse \& Aussem(2016)Gasse and Aussem]{shallowF4}
Maxime Gasse and Alex Aussem.
\newblock F-measure maximization in multi-label classification with
  conditionally independent label subsets.
\newblock \emph{CoRR}, abs/1604.07759, 2016.
\newblock URL \url{http://arxiv.org/abs/1604.07759}.

\bibitem[Good(1952)]{CE}
Irving~J. Good.
\newblock Rational decisions.
\newblock \emph{Journal of the Royal Statistical Society: Series B
  (Methodological)}, 14\penalty0 (1):\penalty0 107--114, Jan 1952.

\bibitem[Google(2021)]{mobileNetV2}
Google.
\newblock Feature vectors of images with {MobileNet V2} (depth multiplier 1.00)
  trained on {ImageNet} ({ILSVRC-2012-CLS}).
\newblock
  \url{https://tfhub.dev/google/imagenet/mobilenet_v2_100_224/feature_vector/4},
  2021.

\bibitem[Grabocka et~al.(2019)Grabocka, Scholz, and
  Schmidt{-}Thieme]{binaryOpt}
Josif Grabocka, Randolf Scholz, and Lars Schmidt{-}Thieme.
\newblock Learning surrogate losses.
\newblock \emph{CoRR}, abs/1905.10108, 2019.
\newblock URL \url{http://arxiv.org/abs/1905.10108}.

\bibitem[Guo et~al.(2017)Guo, Pleiss, Sun, and Weinberger]{predsConf}
Chuan Guo, Geoff Pleiss, Yu~Sun, and Kilian~Q. Weinberger.
\newblock On calibration of modern neural networks.
\newblock In Doina Precup and Yee~Whye Teh (eds.), \emph{Proceedings of the
  34th International Conference on Machine Learning}, volume~70 of
  \emph{Proceedings of Machine Learning Research}, pp.\  1321--1330. PMLR,
  06--11 Aug 2017.

\bibitem[Hanahan \& Weinberg(2011)Hanahan and Weinberg]{cancerHallmarks}
Douglas Hanahan and Robert~A. Weinberg.
\newblock Hallmarks of cancer: The next generation.
\newblock \emph{Cell}, 144\penalty0 (5):\penalty0 646--674, Mar 2011.

\bibitem[He et~al.(2019)He, Wang, Liu, Feng, and Wu]{oldarXiv}
Jun He, Liqun Wang, Liu Liu, Jiao Feng, and Hao Wu.
\newblock Long document classification from local word glimpses via recurrent
  attention learning.
\newblock \emph{IEEE Access}, 7:\penalty0 40707--40718, 2019.

\bibitem[Howard \& Ruder(2018)Howard and Ruder]{ULMFit}
Jeremy Howard and Sebastian Ruder.
\newblock Universal language model fine-tuning for text classification.
\newblock \emph{Proceedings of the 56th Annual Meeting of the Association for
  Computational Linguistics (Volume 1: Long Papers)}, 2018.

\bibitem[Huang et~al.(2015)Huang, Xu, Wang, and Silamu]{sigmoid}
Hao Huang, Haihua Xu, Xianhui Wang, and Wushour Silamu.
\newblock Maximum f1-score discriminative training criterion for automatic
  mispronunciation detection.
\newblock \emph{IEEE/ACM Transactions on Audio, Speech, and Language
  Processing}, 23\penalty0 (4):\penalty0 787--797, Apr 2015.

\bibitem[Huggingface(2021)]{DistilBERTModel}
Huggingface.
\newblock {DistilBERT}.
\newblock \url{https://huggingface.co/transformers/model_doc/distilbert.html},
  2021.

\bibitem[Hull(1994)]{IRClassStat}
David~A Hull.
\newblock \emph{Information retrieval using statistical classification}.
\newblock PhD thesis, Stanford University, 1994.

\bibitem[H{\"u}llermeier et~al.(2008)H{\"u}llermeier, F{\"u}rnkranz, Cheng, and
  Brinker]{pairwiseBinary}
Eyke H{\"u}llermeier, Johannes F{\"u}rnkranz, Weiwei Cheng, and Klaus Brinker.
\newblock Label ranking by learning pairwise preferences.
\newblock \emph{Artificial Intelligence}, 172\penalty0 (16):\penalty0
  1897--1916, 2008.

\bibitem[Jain et~al.(2016)Jain, Prabhu, and Varma]{exteremeMissingApplications}
Himanshu Jain, Yashoteja Prabhu, and Manik Varma.
\newblock Extreme multi-label loss functions for recommendation, tagging,
  ranking; other missing label applications.
\newblock In \emph{Proceedings of the 22nd ACM SIGKDD International Conference
  on Knowledge Discovery and Data Mining}, KDD '16, pp.\  935--944, New York,
  NY, USA, 2016. Association for Computing Machinery.

\bibitem[Jain et~al.(2019)Jain, Balasubramanian, Chunduri, and
  Varma]{extremeMilliionsSlice}
Himanshu Jain, Venkatesh Balasubramanian, Bhanu Chunduri, and Manik Varma.
\newblock Slice: Scalable linear extreme classifiers trained on 100 million
  labels for related searches.
\newblock In \emph{WSDM '19, February 11--15, 2019, Melbourne, VIC, Australia}.
  ACM, February 2019.
\newblock Best Paper Award at WSDM '19.

\bibitem[Jernite et~al.(2017)Jernite, Choromanska, and
  Sontag]{extremeClassification}
Yacine Jernite, Anna Choromanska, and David Sontag.
\newblock Simultaneous learning of trees and representations for extreme
  classification and density estimation.
\newblock In \emph{Proceedings of the 34th International Conference on Machine
  Learning - Volume 70}, ICML'17, pp.\  1665--1674. JMLR.org, 2017.

\bibitem[Joulin et~al.(2017)Joulin, Grave, Bojanowski, and Mikolov]{PAL}
Armand Joulin, Edouard Grave, Piotr Bojanowski, and Tomas Mikolov.
\newblock Bag of tricks for efficient text classification.
\newblock In \emph{Proceedings of the 15th Conference of the {E}uropean Chapter
  of the Association for Computational Linguistics: Volume 2, Short Papers},
  pp.\  427--431, Valencia, Spain, April 2017. Association for Computational
  Linguistics.

\bibitem[Kang \& Kim(2003)Kang and Kim]{queryClassification}
In-Ho Kang and GilChang Kim.
\newblock Query type classification for web document retrieval.
\newblock In \emph{Proceedings of the 26th Annual International ACM SIGIR
  Conference on Research and Development in Informaion Retrieval}, SIGIR '03,
  pp.\  64--71, New York, NY, USA, 2003. Association for Computing Machinery.

\bibitem[Koyejo et~al.(2015)Koyejo, Natarajan, Ravikumar, and
  Dhillon]{multilabelMetrics}
Oluwasanmi~O. Koyejo, Nagarajan Natarajan, Pradeep~K. Ravikumar, and
  Inderjit~S. Dhillon.
\newblock Consistent multilabel classification.
\newblock In \emph{Advances in Neural Information Processing Systems},
  volume~28, pp.\  3321--3329. Curran Associates, Inc., 2015.

\bibitem[Krizhevsky et~al.(2017)Krizhevsky, Sutskever, and Hinton]{saturation}
Alex Krizhevsky, Ilya Sutskever, and Geoffrey~E. Hinton.
\newblock Imagenet classification with deep convolutional neural networks.
\newblock \emph{Communications of the ACM}, 60\penalty0 (6):\penalty0 84--90,
  May 2017.

\bibitem[Lapin et~al.(2015)Lapin, Hein, and Schiele]{topKmulticlassSVM}
Maksim Lapin, Matthias Hein, and Bernt Schiele.
\newblock Top-k multiclass {SVM}.
\newblock In \emph{Proceedings of the 28th International Conference on Neural
  Information Processing Systems - Volume 1}, NIPS'15, pp.\  325--333,
  Cambridge, MA, USA, 2015. MIT Press.

\bibitem[Lapin et~al.(2016)Lapin, Hein, and Schiele]{lossTopKError}
Maksim Lapin, Matthias Hein, and Bernt Schiele.
\newblock Loss functions for top-k error: Analysis and insights.
\newblock \emph{2016 IEEE Conference on Computer Vision and Pattern Recognition
  (CVPR)}, Jun 2016.

\bibitem[Larsson et~al.(2014)Larsson, Silins, Guo, Korhonen, Stenius, and
  Berglund]{chemExposure}
Kristin Larsson, Ilona Silins, Yufan Guo, Anna Korhonen, Ulla Stenius, and
  Marika Berglund.
\newblock Text mining for improved human exposure assessment.
\newblock \emph{Toxicology Letters}, 229:\penalty0 S119, Sep 2014.

\bibitem[Lehmann et~al.(2015)Lehmann, Isele, Jakob, Jentzsch, Kontokostas,
  Mendes, Hellmann, Morsey, Van~Kleef, Auer, and Bizer]{lehmann2015dbpedia}
Jens Lehmann, Robert Isele, Max Jakob, Anja Jentzsch, Dimitris Kontokostas,
  Pablo~N Mendes, Sebastian Hellmann, Mohamed Morsey, Patrick Van~Kleef,
  S{\"o}ren Auer, and Christian Bizer.
\newblock {DBpedia} -- {A} large-scale, multilingual knowledge base extracted
  from {Wikipedia}.
\newblock \emph{Semantic Web}, 6\penalty0 (2):\penalty0 167--195, 2015.

\bibitem[Leng et~al.(2022)Leng, Tan, Liu, Cubuk, Shi, Cheng, and
  Anguelov]{polyloss}
Zhaoqi Leng, Mingxing Tan, Chenxi Liu, Ekin~Dogus Cubuk, Jay Shi, Shuyang
  Cheng, and Dragomir Anguelov.
\newblock {PolyLoss}: A polynomial expansion perspective of classification loss
  functions.
\newblock In \emph{International Conference on Learning Representations}, 2022.

\bibitem[Li et~al.(2017)Li, Song, and Luo]{multitaskLabelImages}
Yuncheng Li, Yale Song, and Jiebo Luo.
\newblock Improving pairwise ranking for multi-label image classification.
\newblock \emph{2017 IEEE Conference on Computer Vision and Pattern Recognition
  (CVPR)}, Jul 2017.

\bibitem[Lin et~al.(2014)Lin, Maire, Belongie, Hays, Perona, Ramanan,
  Doll{\'a}r, and Zitnick]{COCO}
Tsung-Yi Lin, Michael Maire, Serge Belongie, James Hays, Pietro Perona, Deva
  Ramanan, Piotr Doll{\'a}r, and C.~Lawrence Zitnick.
\newblock Microsoft coco: Common objects in context.
\newblock \emph{Lecture Notes in Computer Science}, pp.\  740--755, 2014.

\bibitem[Lin et~al.(2017)Lin, Goyal, Girshick, He, and Dollar]{focalLoss}
Tsung-Yi Lin, Priya Goyal, Ross Girshick, Kaiming He, and Piotr Dollar.
\newblock Focal loss for dense object detection.
\newblock \emph{2017 IEEE International Conference on Computer Vision (ICCV)},
  Oct 2017.

\bibitem[Lipton et~al.(2014)Lipton, Elkan, and Naryanaswamy]{threshForF1}
Zachary~C. Lipton, Charles Elkan, and Balakrishnan Naryanaswamy.
\newblock Optimal thresholding of classifiers to maximize f1 measure.
\newblock \emph{Lecture Notes in Computer Science}, pp.\  225--239, 2014.

\bibitem[Liu et~al.(2017)Liu, Chang, Wu, and Yang]{extremeSIGIR}
Jingzhou Liu, Wei-Cheng Chang, Yuexin Wu, and Yiming Yang.
\newblock Deep learning for extreme multi-label text classification.
\newblock \emph{Proceedings of the 40th International ACM SIGIR Conference on
  Research and Development in Information Retrieval}, Aug 2017.

\bibitem[Liu et~al.(2021)Liu, Zhang, Yang, Su, and Zhu]{Query2Label}
Shilong Liu, Lei Zhang, Xiao Yang, Hang Su, and Jun Zhu.
\newblock Query2label: {A} simple transformer way to multi-label
  classification.
\newblock \emph{CoRR}, abs/2107.10834, 2021.

\bibitem[Loza~Mencia \& Furnkranz(2008)Loza~Mencia and Furnkranz]{pairwiseNet}
Eneldo Loza~Mencia and Johannes Furnkranz.
\newblock Pairwise learning of multilabel classifications with perceptrons.
\newblock In \emph{2008 IEEE International Joint Conference on Neural Networks
  (IEEE World Congress on Computational Intelligence)}, pp.\  2899--2906, 2008.

\bibitem[Madjarov et~al.(2012)Madjarov, Kocev, Gjorgjevikj, and D{\v
  z}eroski]{multilabelMethods}
Gjorgji Madjarov, Dragi Kocev, Dejan Gjorgjevikj, and Sa{\v s}o D{\v z}eroski.
\newblock An extensive experimental comparison of methods for multi-label
  learning.
\newblock \emph{Pattern recognition}, 45\penalty0 (9):\penalty0 3084--3104,
  2012.

\bibitem[Manning et~al.(2008)Manning, Raghavan, and Sch{\"u}tze]{introIR}
Christopher~D. Manning, Prabhakar Raghavan, and Hinrich Sch{\"u}tze.
\newblock \emph{Introduction to Information Retrieval}.
\newblock Cambridge University Press, 2008.

\bibitem[Menon et~al.(2019)Menon, Rawat, Reddi, and Kumar]{multilabelReduction}
Aditya~K Menon, Ankit~Singh Rawat, Sashank Reddi, and Sanjiv Kumar.
\newblock Multilabel reductions: what is my loss optimising?
\newblock In H.~Wallach, H.~Larochelle, A.~Beygelzimer, F.~d\textquotesingle
  Alch\'{e}-Buc, E.~Fox, and R.~Garnett (eds.), \emph{Advances in Neural
  Information Processing Systems}, volume~32. Curran Associates, Inc., 2019.

\bibitem[Milbich et~al.(2020)Milbich, Ghori, Diego, and
  Ommer]{unsupervisedImage}
Timo Milbich, Omair Ghori, Ferran Diego, and Bj{\"o}rn Ommer.
\newblock Unsupervised representation learning by discovering reliable image
  relations.
\newblock \emph{Pattern Recognition}, 102:\penalty0 107107, Jun 2020.

\bibitem[Nakano et~al.(2020)Nakano, Cerri, and Vens]{activeLearningMultiLabel}
Felipe~Kenji Nakano, Ricardo Cerri, and Celine Vens.
\newblock Active learning for hierarchical multi-label classification.
\newblock \emph{Data Mining and Knowledge Discovery}, 34\penalty0 (5):\penalty0
  1496--1530, 2020.

\bibitem[Narasimhan et~al.(2015)Narasimhan, Kar, and Jain]{multiclassOpt1}
Harikrishna Narasimhan, Purushottam Kar, and Prateek Jain.
\newblock Optimizing non-decomposable performance measures: A tale of two
  classes.
\newblock In \emph{Proceedings of the 32nd International Conference on
  International Conference on Machine Learning - Volume 37}, ICML'15, pp.\
  199--208. JMLR.org, 2015.

\bibitem[Neha(2018)]{moviePostersData}
Neha.
\newblock Movie genre from its poster.
\newblock \url{https://www.kaggle.com/neha1703/movie-genre-from-its-poster},
  2018.

\bibitem[Padilla et~al.(2020)Padilla, Netto, and da~Silva]{mAP}
Rafael Padilla, Sergio~L. Netto, and Eduardo A.~B. da~Silva.
\newblock A survey on performance metrics for object-detection algorithms.
\newblock In \emph{2020 International Conference on Systems, Signals and Image
  Processing (IWSSIP)}, pp.\  237--242, 2020.

\bibitem[Paszke et~al.(2019)Paszke, Gross, Massa, Lerer, Bradbury, Chanan,
  Killeen, Lin, Gimelshein, Antiga, Desmaison, Kopf, Yang, DeVito, Raison,
  Tejani, Chilamkurthy, Steiner, Fang, Bai, and Chintala]{pytorch}
Adam Paszke, Sam Gross, Francisco Massa, Adam Lerer, James Bradbury, Gregory
  Chanan, Trevor Killeen, Zeming Lin, Natalia Gimelshein, Luca Antiga, Alban
  Desmaison, Andreas Kopf, Edward Yang, Zachary DeVito, Martin Raison, Alykhan
  Tejani, Sasank Chilamkurthy, Benoit Steiner, Lu~Fang, Junjie Bai, and Soumith
  Chintala.
\newblock Pytorch: An imperative style, high-performance deep learning library.
\newblock In H.~Wallach, H.~Larochelle, A.~Beygelzimer, F.~d\textquotesingle
  Alch\'{e}-Buc, E.~Fox, and R.~Garnett (eds.), \emph{Advances in Neural
  Information Processing Systems 32}, pp.\  8024--8035. Curran Associates,
  Inc., 2019.
\newblock URL
  \url{http://papers.neurips.cc/paper/9015-pytorch-an-imperative-style-high-performance-deep-learning-library.pdf}.

\bibitem[Patel et~al.(2022)Patel, Tolias, and Matas]{recallLoss}
Yash Patel, Giorgos Tolias, and Jiri Matas.
\newblock Recall@k surrogate loss with large batches and similarity mixup.
\newblock \emph{CVPR}, 2022.

\bibitem[Perotte et~al.(2014)Perotte, Pivovarov, Natarajan, Weiskopf, Wood, and
  Elhadad]{diagnosisCode}
Adler Perotte, Rimma Pivovarov, Karthik Natarajan, Nicole Weiskopf, Frank Wood,
  and No{\'e}mie Elhadad.
\newblock Diagnosis code assignment: Models and evaluation metrics.
\newblock \emph{Journal of the American Medical Informatics Association},
  21\penalty0 (2):\penalty0 231--237, Mar 2014.

\bibitem[Prabhu et~al.(2018)Prabhu, Kag, Harsola, Agrawal, and
  Varma]{extremeParabel}
Yashoteja Prabhu, Anil Kag, Shrutendra Harsola, Rahul Agrawal, and Manik Varma.
\newblock Parabel: Partitioned label trees for extreme classification with
  application to dynamic search advertising.
\newblock In \emph{Proceedings of the 2018 World Wide Web Conference}, WWW '18,
  pp.\  993--1002, Republic and Canton of Geneva, CHE, 2018. International
  World Wide Web Conferences Steering Committee.

\bibitem[Ramaswamy et~al.(2014)Ramaswamy, Babu, Agarwal, and
  Williamson]{multiclassToBinary3}
Harish~G. Ramaswamy, Balaji~Srinivasan Babu, Shivani Agarwal, and Robert~C.
  Williamson.
\newblock On the consistency of output code based learning algorithms for
  multiclass learning problems.
\newblock In Maria~Florina Balcan, Vitaly Feldman, and Csaba Szepesv{\'a}ri
  (eds.), \emph{Proceedings of The 27th Conference on Learning Theory},
  volume~35 of \emph{Proceedings of Machine Learning Research}, pp.\  885--902,
  Barcelona, Spain, 13--15 Jun 2014. PMLR.

\bibitem[Reddi et~al.(2019)Reddi, Kale, Yu, Holtmann-Rice, Chen, and
  Kumar]{stochasticNegativeMining}
Sashank~J. Reddi, Satyen Kale, Felix Yu, Daniel Holtmann-Rice, Jiecao Chen, and
  Sanjiv Kumar.
\newblock Stochastic negative mining for learning with large output spaces.
\newblock In Kamalika Chaudhuri and Masashi Sugiyama (eds.), \emph{Proceedings
  of the Twenty-Second International Conference on Artificial Intelligence and
  Statistics}, volume~89 of \emph{Proceedings of Machine Learning Research},
  pp.\  1940--1949. PMLR, 16--18 Apr 2019.

\bibitem[Ridnik et~al.(2021)Ridnik, Lawen, Noy, Ben~Baruch, Sharir, and
  Friedman]{TresNet}
Tal Ridnik, Hussam Lawen, Asaf Noy, Emanuel Ben~Baruch, Gilad Sharir, and
  Itamar Friedman.
\newblock Tresnet: High performance gpu-dedicated architecture.
\newblock In \emph{Proceedings of the IEEE/CVF Winter Conference on
  Applications of Computer Vision (WACV)}, pp.\  1400--1409, January 2021.

\bibitem[Rota~Bul\`o et~al.(2018)Rota~Bul\`o, Porzi, and
  Kontschieder]{inplaceABN}
Samuel Rota~Bul\`o, Lorenzo Porzi, and Peter Kontschieder.
\newblock In-place activated batchnorm for memory-optimized training of dnns.
\newblock In \emph{Proceedings of the IEEE Conference on Computer Vision and
  Pattern Recognition}, 2018.

\bibitem[Sandler et~al.(2018)Sandler, Howard, Zhu, Zhmoginov, and
  Chen]{mobileNet}
Mark Sandler, Andrew Howard, Menglong Zhu, Andrey Zhmoginov, and Liang-Chieh
  Chen.
\newblock {MobileNetV2}: Inverted residuals and linear bottlenecks.
\newblock In \emph{2018 IEEE/CVF Conference on Computer Vision and Pattern
  Recognition (CVPR)}, pp.\  4510--4520, 06 2018.

\bibitem[Sanh et~al.(2019)Sanh, Debut, Chaumond, and Wolf]{distilBert}
Victor Sanh, Lysandre Debut, Julien Chaumond, and Thomas Wolf.
\newblock {DistilBERT}, a distilled version of {BERT:} smaller, faster, cheaper
  and lighter.
\newblock \emph{CoRR}, abs/1910.01108, 2019.

\bibitem[Sanyal et~al.(2018)Sanyal, Kumar, Kar, Chawla, and
  Sebastiani]{multiclassOptDL}
Amartya Sanyal, Pawan Kumar, Purushottam Kar, Sanjay Chawla, and Fabrizio
  Sebastiani.
\newblock Optimizing non-decomposable measures with deep networks.
\newblock \emph{Machine Learning}, 107, 09 2018.
\newblock \doi{10.1007/s10994-018-5736-y}.

\bibitem[Shen et~al.(2017)Shen, Mu, Yang, Liu, Liu, Song, and
  Shen]{imageClassification}
Fumin Shen, Yadong Mu, Yang Yang, Wei Liu, Li~Liu, Jingkuan Song, and Heng~Tao
  Shen.
\newblock Classification by retrieval: Binarizing data and classifiers.
\newblock In \emph{Proceedings of the 40th International ACM SIGIR Conference
  on Research and Development in Information Retrieval}, SIGIR '17, pp.\
  595--604, New York, NY, USA, 2017. Association for Computing Machinery.

\bibitem[Smith et~al.(2017)Smith, Kindermans, and Le]{bigBSArxiv}
Samuel~L. Smith, Pieter{-}Jan Kindermans, and Quoc~V. Le.
\newblock Don't decay the learning rate, increase the batch size.
\newblock \emph{CoRR}, abs/1711.00489, 2017.

\bibitem[Soleimani \& Miller(2017)Soleimani and Miller]{multiInstance}
Hossein Soleimani and David~J. Miller.
\newblock Semisupervised, multilabel, multi-instance learning for structured
  data.
\newblock \emph{Neural Computation}, 29\penalty0 (4):\penalty0 1053--1102,
  2017.

\bibitem[Tewari \& Bartlett(2005)Tewari and Bartlett]{multiclassToBinary2}
Ambuj Tewari and Peter~L. Bartlett.
\newblock On the consistency of multiclass classification methods.
\newblock In Peter Auer and Ron Meir (eds.), \emph{Learning Theory}, pp.\
  143--157, Berlin, Heidelberg, 2005. Springer Berlin Heidelberg.

\bibitem[Tsoumakas \& Katakis(2007)Tsoumakas and Katakis]{hammingLoss}
Grigorios Tsoumakas and Ioannis Katakis.
\newblock Multi-label classification: An overview.
\newblock \emph{International Journal of Data Warehousing and Mining (IJDWM)},
  3\penalty0 (3):\penalty0 1--13, 2007.

\bibitem[Tukey(1977)]{tukey}
John~Wilder Tukey.
\newblock \emph{Exploratory data analysis}.
\newblock Addison-Wesley series in behavioral science: quantitative methods.
  Addison-Wesley, Reading, Mass., 1977.

\bibitem[Waegeman et~al.(2014)Waegeman, Dembczy\'{n}ki, Jachnik, Cheng, and
  H\"{u}llermeier]{F-inference}
Willem Waegeman, Krzysztof Dembczy\'{n}ki, Arkadiusz Jachnik, Weiwei Cheng, and
  Eyke H\"{u}llermeier.
\newblock On the bayes-optimality of f-measure maximizers.
\newblock \emph{J. Mach. Learn. Res.}, 15\penalty0 (1):\penalty0 3333--3388,
  January 2014.

\bibitem[Wang et~al.(2016)Wang, Yang, Mao, Huang, Huang, and
  Xu]{multilabelImage1}
Jiang Wang, Yi~Yang, Junhua Mao, Zhiheng Huang, Chang Huang, and Wei Xu.
\newblock Cnn-rnn: A unified framework for multi-label image classification.
\newblock In \emph{Proceedings of the IEEE Conference on Computer Vision and
  Pattern Recognition (CVPR)}, June 2016.

\bibitem[Wang et~al.(2020)Wang, Sun, Cheng, Jiang, Deng, Zhao, Liu, Mu, Tan,
  Wang, and et~al.]{highResRepresentation}
Jingdong Wang, Ke~Sun, Tianheng Cheng, Borui Jiang, Chaorui Deng, Yang Zhao,
  Dong Liu, Yadong Mu, Mingkui Tan, Xinggang Wang, and et~al.
\newblock Deep high-resolution representation learning for visual recognition.
\newblock \emph{IEEE Transactions on Pattern Analysis and Machine
  Intelligence}, pp.\  1--1, 2020.

\bibitem[Wang et~al.(2018)Wang, Li, Golbandi, Bendersky, and
  Najork]{lambdaLoss}
Xuanhui Wang, Cheng Li, Nadav Golbandi, Mike Bendersky, and Marc Najork.
\newblock The {LambdaLoss} framework for ranking metric optimization.
\newblock In \emph{Proceedings of The 27th ACM International Conference on
  Information and Knowledge Management (CIKM '18)}, pp.\  1313--1322, 2018.

\bibitem[Wei et~al.(2016)Wei, Xia, Lin, Huang, Ni, Dong, Zhao, and
  Yan]{multilabelImage2}
Yunchao Wei, Wei Xia, Min Lin, Junshi Huang, Bingbing Ni, Jian Dong, Yao Zhao,
  and Shuicheng Yan.
\newblock Hcp: A flexible cnn framework for multi-label image classification.
\newblock \emph{IEEE Transactions on Pattern Analysis and Machine
  Intelligence}, 38\penalty0 (9):\penalty0 1901--1907, 2016.

\bibitem[Wu et~al.(2019)Wu, Chen, Fan, Zhang, Hou, Liu, and Zhang]{tencent}
Baoyuan Wu, Weidong Chen, Yanbo Fan, Yong Zhang, Jinlong Hou, Jie Liu, and Tong
  Zhang.
\newblock Tencent ml-images: A large-scale multi-label image database for
  visual representation learning.
\newblock \emph{IEEE Access}, 7:\penalty0 172683--172693, 2019.

\bibitem[Wu \& Zhou(2017)Wu and Zhou]{unifiedView}
Xi-Zhu Wu and Zhi-Hua Zhou.
\newblock A unified view of multi-label performance measures.
\newblock In Doina Precup and Yee~Whye Teh (eds.), \emph{Proceedings of the
  34th International Conference on Machine Learning}, volume~70 of
  \emph{Proceedings of Machine Learning Research}, pp.\  3780--3788. PMLR,
  06--11 Aug 2017.

\bibitem[Wydmuch et~al.(2018)Wydmuch, Jasinska, Kuznetsov, Busa-Fekete, and
  Dembczy\'{n}ski]{OVATheory}
Marek Wydmuch, Kalina Jasinska, Mikhail Kuznetsov, R\'{o}bert Busa-Fekete, and
  Krzysztof Dembczy\'{n}ski.
\newblock A no-regret generalization of hierarchical softmax to extreme
  multi-label classification.
\newblock In \emph{Proceedings of the 32nd International Conference on Neural
  Information Processing Systems}, NIPS'18, pp.\  6358--6368, Red Hook, NY,
  USA, 2018. Curran Associates Inc.

\bibitem[Xiao et~al.(2010)Xiao, Hays, Ehinger, Oliva, and
  Torralba]{faceDetection}
Jianxiong Xiao, James Hays, Krista~A. Ehinger, Aude Oliva, and Antonio
  Torralba.
\newblock Sun database: Large-scale scene recognition from abbey to zoo.
\newblock In \emph{2010 IEEE Computer Society Conference on Computer Vision and
  Pattern Recognition}, pp.\  3485--3492, 2010.

\bibitem[Yang(2004)]{statTextCategorization}
Yiming Yang.
\newblock An evaluation of statistical approaches to text categorization.
\newblock \emph{Information Retrieval}, 1:\penalty0 69--90, 2004.

\bibitem[Yang et~al.(2019)Yang, Dai, Yang, Carbonell, Salakhutdinov, and
  Le]{XLNet}
Zhilin Yang, Zihang Dai, Yiming Yang, Jaime Carbonell, Russ~R Salakhutdinov,
  and Quoc~V Le.
\newblock Xlnet: Generalized autoregressive pretraining for language
  understanding.
\newblock In H.~Wallach, H.~Larochelle, A.~Beygelzimer, F.~d\textquotesingle
  Alch\'{e}-Buc, E.~Fox, and R.~Garnett (eds.), \emph{Advances in Neural
  Information Processing Systems}, volume~32, pp.\  5753--5763. Curran
  Associates, Inc., 2019.

\bibitem[Ye et~al.(2012)Ye, Chai, Lee, and Chieu]{shallowF1}
Nan Ye, Kian Ming~Adam Chai, Wee~Sun Lee, and Hai~Leong Chieu.
\newblock Optimizing f-measure: A tale of two approaches.
\newblock In \emph{ICML}, 2012.
\newblock URL \url{http://icml.cc/2012/papers/175.pdf}.

\bibitem[Yen et~al.(2017)Yen, Huang, Dai, Ravikumar, Dhillon, and
  Xing]{extremePPDsparse}
Ian~E.H. Yen, Xiangru Huang, Wei Dai, Pradeep Ravikumar, Inderjit Dhillon, and
  Eric Xing.
\newblock Ppdsparse: A parallel primal-dual sparse method for extreme
  classification.
\newblock In \emph{Proceedings of the 23rd ACM SIGKDD International Conference
  on Knowledge Discovery and Data Mining}, KDD '17, pp.\  545--553, New York,
  NY, USA, 2017. Association for Computing Machinery.

\bibitem[Yu et~al.(2014)Yu, Jain, Kar, and Dhillon]{exteremeMissing}
Hsiang-Fu Yu, Prateek Jain, Purushottam Kar, and Inderjit~S. Dhillon.
\newblock Large-scale multi-label learning with missing labels.
\newblock In \emph{Proceedings of the 31st International Conference on
  International Conference on Machine Learning - Volume 32}, ICML'14, pp.\
  I--593--I--601. JMLR.org, 2014.

\bibitem[Zaheer et~al.(2020)Zaheer, Guruganesh, Dubey, Ainslie, Alberti,
  Ontanon, Pham, Ravula, Wang, Yang, and Ahmed]{bigBird}
Manzil Zaheer, Guru Guruganesh, Kumar~Avinava Dubey, Joshua Ainslie, Chris
  Alberti, Santiago Ontanon, Philip Pham, Anirudh Ravula, Qifan Wang, Li~Yang,
  and Amr Ahmed.
\newblock Big bird: Transformers for longer sequences.
\newblock In \emph{Advances in Neural Information Processing Systems}, 2020.

\bibitem[Zhang \& Zhou(2006)Zhang and Zhou]{multilabelBackprop}
Min-Ling Zhang and Zhi-Hua Zhou.
\newblock Multilabel neural networks with applications to functional genomics
  and text categorization.
\newblock \emph{IEEE Transactions on Knowledge and Data Engineering},
  18\penalty0 (10):\penalty0 1338--1351, Oct 2006.

\bibitem[Zhang \& Zhou(2007)Zhang and Zhou]{ML-KNN}
Min-Ling Zhang and Zhi-Hua Zhou.
\newblock Ml-knn: A lazy learning approach to multi-label learning.
\newblock \emph{Pattern Recognition}, 40\penalty0 (7):\penalty0 2038--2048, Jul
  2007.

\bibitem[Zhang \& Zhou(2014)Zhang and Zhou]{multilabelReview}
Min-Ling Zhang and Zhi-Hua Zhou.
\newblock A review on multi-label learning algorithms.
\newblock \emph{IEEE Transactions on Knowledge and Data Engineering},
  26\penalty0 (8):\penalty0 1819--1837, Aug 2014.

\bibitem[Zhang et~al.(2020)Zhang, Ramaswamy, and Agarwal]{shallowF5}
Mingyuan Zhang, Harish~G. Ramaswamy, and Shivani Agarwal.
\newblock Convex calibrated surrogates for the multi-label f-measure.
\newblock In \emph{Proceedings of the 37th International Conference on Machine
  Learning}, ICML'20. JMLR.org, 2020.

\bibitem[Zhang(2004)]{multiclassToBinary1}
Tong Zhang.
\newblock {Statistical behavior and consistency of classification methods based
  on convex risk minimization}.
\newblock \emph{The Annals of Statistics}, 32\penalty0 (1):\penalty0 56 -- 85,
  2004.

\bibitem[Zhang et~al.(2015)Zhang, Zhao, and LeCun]{textClassificationLeCun}
Xiang Zhang, Junbo Zhao, and Yann LeCun.
\newblock Character-level convolutional networks for text classification.
\newblock In \emph{Proceedings of the 28th International Conference on Neural
  Information Processing Systems - Volume 1}, NIPS'15, pp.\  649--657,
  Cambridge, MA, USA, 2015. MIT Press.

\bibitem[Zhou et~al.(2012)Zhou, Zhang, Huang, and Li]{multiInstanceMultiLabel}
Zhi-Hua Zhou, Min-Ling Zhang, Sheng-Jun Huang, and Yu-Feng Li.
\newblock Multi-instance multi-label learning.
\newblock \emph{Artificial Intelligence}, 176\penalty0 (1):\penalty0 2291 --
  2320, 2012.

\bibitem[Zhu et~al.(2017)Zhu, Li, Ouyang, Yu, and Wang]{multilabelImage3}
Feng Zhu, Hongsheng Li, Wanli Ouyang, Nenghai Yu, and Xiaogang Wang.
\newblock Learning spatial regularization with image-level supervisions for
  multi-label image classification.
\newblock In \emph{Proceedings of the IEEE Conference on Computer Vision and
  Pattern Recognition (CVPR)}, July 2017.

\end{thebibliography}
\bibliographystyle{tmlr-style-file-main/tmlr}

\begin{appendices}

\section{Evaluation Metrics}
\label{sec:evalMetrics}

In our experimental evaluation, we consider a suite of metrics that are commonly used in the evaluation of multilabel classification to measure the effectiveness of multilabel prediction. These metrics are based on the confusion matrix and for which we provided smoothed surrogates to optimize directly (see Table \ref{tab:confusion-matrix}).

\begin{table*}[htbp]
\centering
\caption{Confusion matrix with our proposed smoothed confusion matrix entries, $\widetilde{\mathit{tp}}$, $\widetilde{\mathit{fp}}$, $\widetilde{\mathit{fn}}$ and $\widetilde{\mathit{tn}}$ and six derived loss functions that use these smoothed confusion matrix entries. $\mathcal{L}_{\widetilde{\mathit{F1}}}$ is used in our experiments.}
\label{tab:confusion-matrix}
\def\arraystretch{1.1}
\begin{tabular}{ccc c} 
\toprule
 & \textbf{Condition} & \textbf{Condition} & \multirow{2}{*}{$\mathcal{L}_{\widetilde{\mathit{Accuracy}}}= \frac{\widetilde{\mathit{tp}} + \widetilde{\mathit{tn}}}{\widetilde{\mathit{tp}} + \widetilde{\mathit{fp}} + \widetilde{\mathit{tn}} + \widetilde{\mathit{fn}}}$} \\
 & \textbf{positive} &  \textbf{negative} & 
\\ 
\midrule
\textbf{~Predicted~} & True positive & False positive & \multirow{2}{*}{$\mathcal{L}_{\widetilde{\mathit{Precision}}}= \frac{\widetilde{\mathit{tp}}}{\widetilde{\mathit{tp}} + \widetilde{\mathit{fp}}}$} 
\\
\textbf{positive} & $\widetilde{\mathit{tp}}=\sum \mathbf{S}(\hat{\mathbf{y}}) \odot \mathbf{y}$ & $\widetilde{\mathit{fp}}= \sum \mathbf{S}(\hat{\mathbf{y}}) \odot (\mathds{1} - \mathbf{y})$ & 
\\ 
\hline
\textbf{Predicted} & False negative & True Negative & \multirow{2}{*}{$\mathcal{L}_{\widetilde{\mathit{NPV}}}= \frac{\widetilde{\mathit{tn}}}{\widetilde{\mathit{tn}} + \widetilde{\mathit{fn}}}$} 
\\
\textbf{negative} & $\widetilde{\mathit{fn}}= \sum (\mathds{1} - \mathbf{S}(\hat{\mathbf{y}})) \odot \mathbf{y}$ & $\widetilde{\mathit{tn}}= \sum (\mathds{1} - \mathbf{S}(\hat{\mathbf{y}})) \odot (\mathds{1} - \mathbf{y})$ & 
\\ 
\midrule
 & \multirow{2}{*}{\hspace{1.2em}$\mathcal{L}_{\widetilde{\mathit{Recall}}}= \frac{\widetilde{\mathit{tp}}}{\widetilde{\mathit{tp}} + \widetilde{\mathit{fn}}}$\hspace{1.2em}}& \multirow{2}{*}{$\mathcal{L}_{\widetilde{\mathit{Specificity}}}= \frac{\widetilde{\mathit{tn}}}{\widetilde{\mathit{fp}} + \widetilde{\mathit{tn}}}$} & \multirow{2}{*}{$\mathcal{L}_{\widetilde{\mathit{F1}}}= \frac{2 \widetilde{\mathit{tp}}}{2 \widetilde{\mathit{tp}}+ \widetilde{\mathit{fn}}+ \widetilde{\mathit{fp}}}$} \\
 & & & \\
\bottomrule
\end{tabular}%
\end{table*}

When true positives and false positives are used, recall that \(t p= \mathds{1}_{\hat{\mathbf{y}} \geq t} \odot \mathbf{y}\) and \(f p= \mathds{1}_{\hat{\mathbf{y}} \geq t} \odot (\mathds{1} - \mathbf{y})\), and thus a threshold \(t\) must be set. For Pascal-VOC and MS-COCO, we set \(t = 0.5\), as is commonly done in the early literature~\citep{multilabelReview, ML-DT}. In the recent literature, the chosen threshold at inference time can vary but was not found to be justified, we thus decide on neutral thresholds before training.

Extending \(F_1\) to multiclass binary classification means deciding whether to pool classes.
In a first pooled iteration, macro \(F_1\)~\citep{multilabelMetrics} equates to creating a single 2x2 confusion matrix for all classes:
\begin{equation}
F_1^{macro} = \frac{\sum^C 2 tp_j}{2 \sum^C tp_j + \sum^C fn_j + \sum^C fp_j},
\end{equation}
with $\sum^C (\cdot)$ as a short form of $\sum_{j=1}^C(\cdot)$, when summing over each class up to the $C$ classes.
Micro \(F_1\) \citep{threshForF1, multilabelMetrics} amounts to creating one confusion matrix per class or unpooling:
\begin{equation}
F_1^{micro} =  \frac{1}{C} \sum_{j=1}^C \frac{2 tp_j}{2 tp_j + fn_j + fp_j} =  \frac{1}{C} \sum_{j=1}^C F_1^j.
\end{equation}
Weighted micro \(F_1\)~\citep{weightedMetrics} is similar but includes weighing to account for class imbalance, i.e., weighing each class by the number of ground truth positives:
\begin{equation}
F_1^{weighted} = \frac{1}{C} \sum_{j=1}^C p_j F_1^j \quad \text{, where } p_j = \sum_i \mathds{1}_{\mathbf{y_i^j} = 1}.
\end{equation}
We also define micro precision%
\begin{equation}
\begin{aligned} P^{micro} &=  \frac{1}{C} \sum_{j=1}^C \frac{tp_j}{tp_j+fp_j}.%
\end{aligned}
\end{equation}
mean Average Precision (mAP) has different definitions. We use mAP as defined for the MS-COCO and Pascal-VOC datasets~\citep{mAP}. Traditionally, Precision and Recall is computed over the intersection of object detection boxes. We use a slightly modified mAP (e.g., in~\citep{ASL}), where precision and recall are computed over the predictions of labels on the whole image. We first obtain the average precision over each class:
\begin{equation}
  \begin{aligned}
    \mathrm{AP}_{\text {all }} &= \sum_{i}\left(R_{i+1}-R_{i}\right) P_{\text {interp }}\left(R_{i+1}\right) \\ P_{\text {interp }}\left(R_{i+1}\right) &= \max _{\tilde{R}: \tilde{R} \geq R_{i+1}} P(\tilde{R}),
  \end{aligned}
\end{equation}
and then compute mean Average Precision:
\begin{equation}
\mathrm{mAP}^{micro}=\frac{1}{C} \sum_{j=1}^{C} \mathrm{AP}_{i}.
\end{equation}
We write micro here to be explicit, but it seems to be mostly computed at the micro level in the literature.

\section{Focal Loss definition}
\label{ap:focalLoss}

We write down the \emph{focalLoss}~\citep{focalLoss}, as it deals specifically with class imbalance and is used as a baseline due to its popularity in the multiclass domain.
\begin{equation}
  \mathcal{L}_{FL} = -\alpha^{\mathrm{j}}\left(1-\hat{y}^{\mathrm{j}}\right)^{\gamma} \log \left(\hat{y}^{\mathrm{j}}\right),
\end{equation}
with $\alpha^j$ and $\gamma$ hyperparameters. In the next section, we further specify the setup for focal loss and cross entropy as benchmarks for \emph{unboundedF1} and \emph{sigmoidF1}.

\section{Compute Time}
\label{app:compute-time}

Table~\ref{table:compute-time} shows compute times in minutes for different on losses and different datasets on a single GPU \emph{g4dn.12xlarge} AWS instance. The run-time is not particularly long, given that we freeze model weights of the pretrained image / text model. 

\begin{table}[h!]
\caption{Average training time over 5 seeds in minutes (60 epochs for MS-COCO and Pascal-VOC, 100 epochs for the reminder two).}
\label{table:compute-time}
\centering
\begin{tabular}{l rrrr}
\toprule
& MS-COCO & Pascal-VOC & arXiv2020 &  moviePosters \\
\midrule
$\mathcal{L}_{\text {BCE}}$[\citeyear{Fisher}] & 856 & 112 &   341  &   58 \\
$\mathcal{L}_{\text {FL}}$[\citeyear{focalLoss}] & 851 &  108 &   428 &   59\\
$\mathcal{L}_{\text {ASL}}$[\citeyear{ASL}] & 856 & 109 &   427 &   59  \\
$\mathcal{L}_{\overline{\mathit{F1}}}$[ours] & 858 &  116 &   381 &   58  \\
$\mathcal{L}_{\widetilde{\mathit{F1}}}$[ours] & 858 & 111 &    351 &   52  \\
\bottomrule
\end{tabular}
\end{table}

\clearpage
\section{Experimental Setup Details}
\label{app:experimental-setup}

\header{moviePosters} consists of images of movie posters and their genres (e.g., \emph{action}, \emph{comedy})~\citep{moviePosters}.\footnote{Labels at \url{https://tinyurl.com/y7ydyedu} and images at \url{https://tinyurl.com/y7lfpvlx}.} The posters and labels have been extracted from IMDB and the dataset was previously used for per-class, post-training thresholding (see Section~\ref{sec:org2aceb9f}). The genre labels in this dataset are not mutually exclusive and of varying counts per movie.

\header{arXiv2020} is a subset of the newly created \emph{arXiv dataset}\footnote{Available at \url{https://tinyurl.com/5kypspya}} with over 1.7 million open source articles and their metadata. Our experiments use the abstracts and categories that are suitably non-mutually exclusive and of varying counts per example. The limited number of labeled classes render the older dataset unsuitable for our experiments.  We write arXiv2020 for the subset of the \emph{arXiv dataset} that only contains documents published in 2020. This results in around 26k documents. There is a longer history of using arXiv to create research datasets; the dataset we use is not to be confused with an earlier long document dataset that only features 11 classes~\citep{oldarXiv}, and was used in a recent long transformer publication~\citep{bigBird}.

\header{pascal-VOC and MS-COCO} stand for Pascal Visual Object Classes Challenge (VOC 2007)~\citep{pascalVOC} and Microsoft Common Objects in Context~\citep{COCO}, respectively. They are object recognition/segmentation datasets. The earlier Pascal-VOC dataset has 20 possible object classes and around 10K examples. The later MS-COCO dataset has 80 possible object classes and around 200K class-annotated examples. Some multilabel classification literature for the image domain use object detection / segmentation datasets to perform multilabel classification:\footnote{See \url{https://paperswithcode.com/task/multi-label-classification}} MS-COCO, Pascal-VOC, NUS-WIDE, etc. (note that transformer models, which effectively distinguish the original objects on the image while predicting labels, perform better on this task~\citep{Query2Label}). Regarding Tresnet-m-21k~\citep{TresNet}, while an L and an XL version of the model exist, the code available online did not allow for correct loading of the weights.

We choose to ignore classes that are underrepresented, in order to give the model a fair chance at learning from at least a few examples. We define underrepresentation as a global irrelevance threshold $b$ for classes: any class $c$ that is represented less than $b$ times is considered irrelevant. We decided to set an irrelevance threshold $b$ on all datasets prior to conducting experiments, so as to not fine-tune for that feature. It was set to 1000 for both \emph{arXiv2020} (145 of the original 155 classes remaining) and \emph{moviePosters} (14 of the 28 classes remaining) and at 10 for \emph{chemicalExposure} (all 38 classes remaining) and \emph{cancerHallmarks} (all 33 classes remaining), in proportion to the number of classes and labels in each dataset. We used all classes for Pascal-VOC and MS-COCO since we are comparing with benchmarks that also do so.

\header{Hyperparameters.} For Pascal-VOC, we found $\{\beta = -0.75; \eta = 10.25\}$ to work best on weightedF1. Given the similarity of the two datasets and the potentially resource-hungry hyperparameter tuning of MS-COCO, we used the same hyperparameters for MS-COCO. For arXiv2020 and moviePosters, $\{\beta = 1; \eta = 9\}$ works best on weightedF1. These hyperparameters where tuned on the validation set and we report on the held out test set. It would be hard to give a general recommendation of hyperparameters, but it seems that $\{\beta = -0.75; \eta = 10.25\}$ is a good basis for image and that $\{\beta = 1; \eta = 9\}$ is a good basis for text.

\paragraph{Setup.} We performed our experiments on Amazon Web Services cloud machines with data parallelization on up to 4 GPUs \textit{g4dn.12xlarge}, with TensorFlow 2~\citep{tensorflow} and PyTorch~\citep{pytorch} as a gradient-descent backend.

\clearpage
\section{Extended Results}
\label{app:extended-results}

\begin{figure}[H]
\centering
\includegraphics[width=\linewidth]{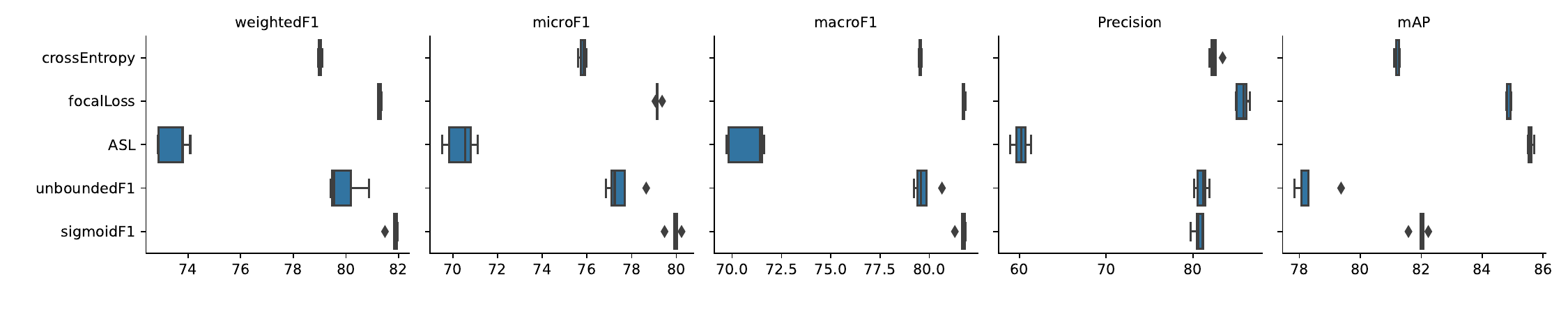}
\vspace{.1\baselineskip}
\caption{Tresnetm21K (CNN) on MS-COCO @0.5.}
\label{fig:bxpltCOCO}
\end{figure}

\begin{figure}[H]
\centering
\includegraphics[width=\linewidth]{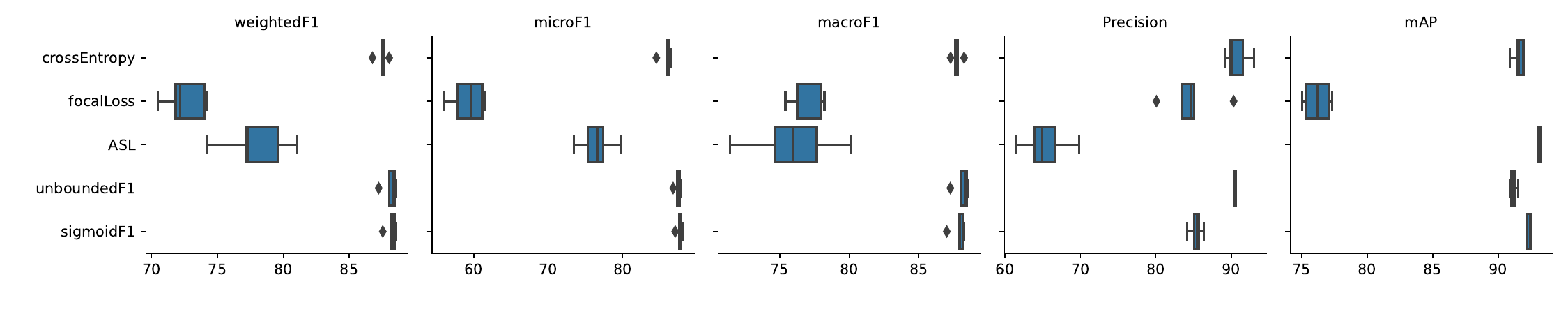}
\vspace{.1\baselineskip}
\caption{Tresnetm21K (CNN) on Pascal-VOC @0.5 (one outlier (<40) for unboundedF1 on each metric ignored for better visualization).}
\label{fig:bxpltPascal}
\end{figure}

\begin{figure}[H]
\centering
\includegraphics[width=\linewidth]{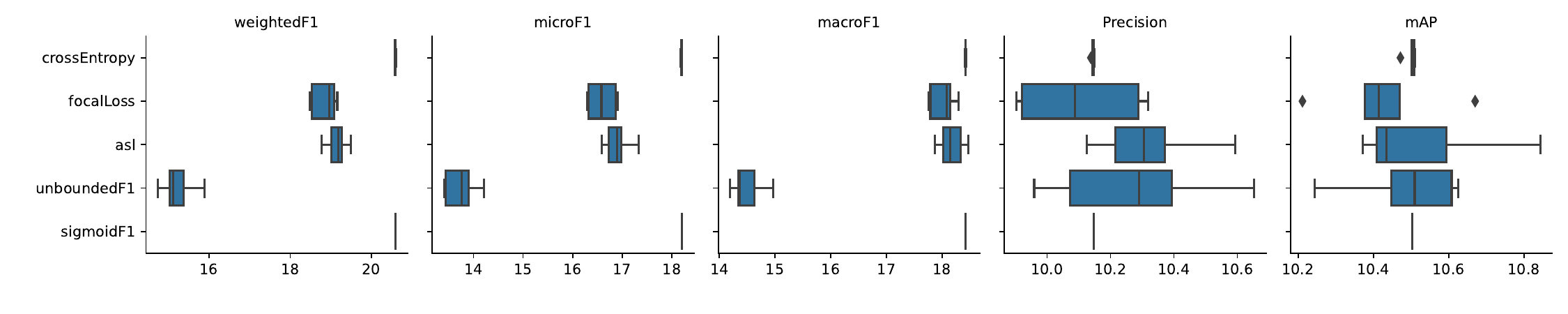}
\vspace{.1\baselineskip}
\caption{DistilBERT (NLP) on arXiv2020 @0.05.}
\label{fig:bxpltArxiv}
\end{figure}

\begin{figure}[H]
\centering
\includegraphics[width=\linewidth]{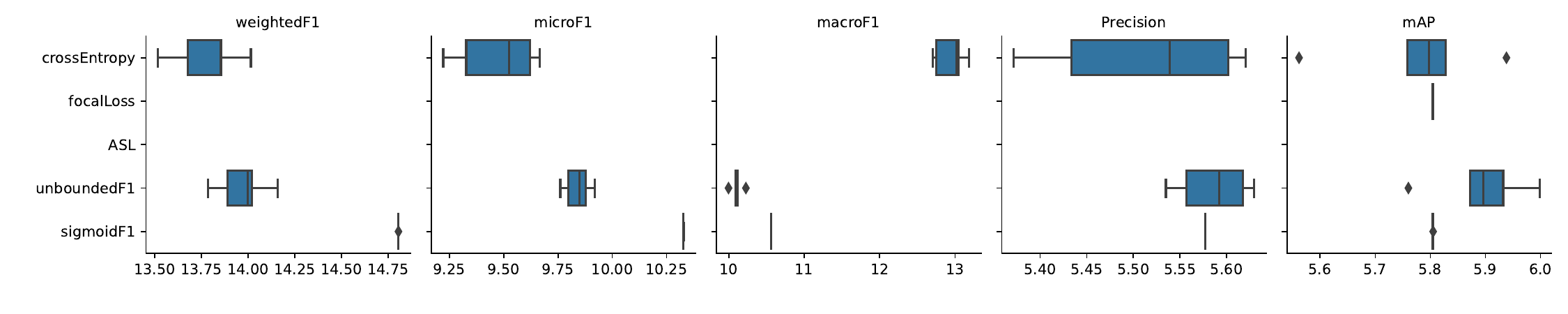}
\vspace{.1\baselineskip}
\caption{MobileNetV2 (CNN) on moviePosters @0.05 (zero values for focalLoss and ASL ignored for better visualization).}
\label{fig:bxpltMovie}
\end{figure}

\begin{figure}[H]
\centering
\includegraphics[width=\linewidth]{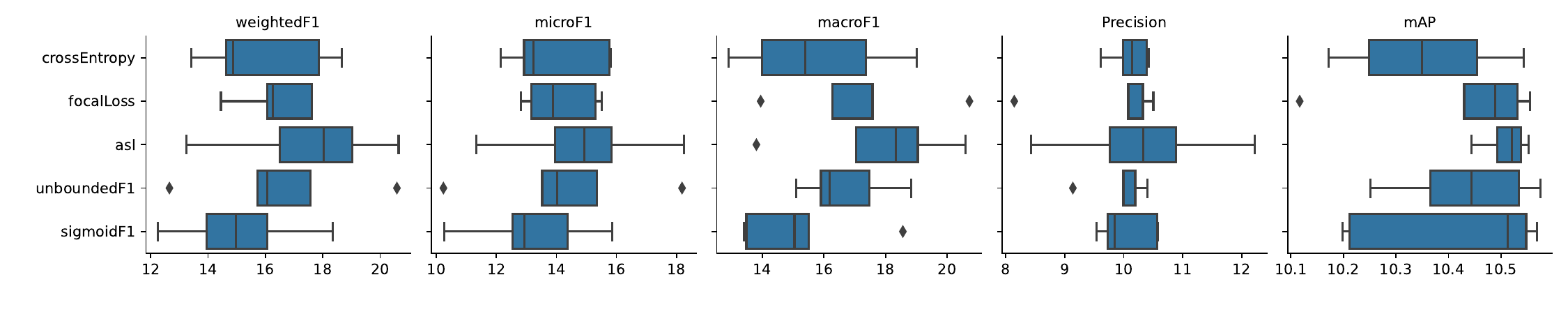}
\vspace{.1\baselineskip}
\caption{DistilBERT (NLP) on arXiv2020 – Threshold moving.}
\label{fig:bxpltArxivAvgThresh}
\end{figure}

Table~\ref{table:overallresults} shows our results as point estimates over 5 training random seeds. This section contains the distributional counterpart of Table~\ref{table:overallresults}, namely boxplots (Figure~\ref{fig:bxpltCOCO}, \ref{fig:bxpltPascal}, \ref{fig:bxpltArxiv} and~\ref{fig:bxpltMovie}) with median and inter quartile range in the blue box. Figure~\ref{fig:bxpltArxivAvgThresh} is the distributional counterpart of Table~\ref{table:avgThresh} (threshold-moving technique on the arXiv dataset) and outlines less conclusive results than for fixed thresholds.

\section{Additional Experiments}
\label{app:addtional-experiments}

This section details additional experiments on two further text datasets from the medical domain. Given that they are relatively small compared to our other benchmark datasets, we keep this discussion in the appendix. Table~\ref{table:allDbatasets} illustrates the difference between our 4 main paper datasets and the 2 appendix datasets. Results on the latter are displayed in Tables~\ref{tab:cancerHallmarks} and~\ref{tab:chemicalExposure}.

ML-NET~\citep{multitaskLabel} has an interesting multitask approach to \emph{fit-algorithm-to-data} methods for multilabel learning with unknown label count on text. The cancerHallmark~\citep{cancerHallmarks}\footnote{Available at \url{https://github.com/sb895/Hallmarks-of-Cancer}} and chemicalExposure~\citep{chemExposure}\footnote{Available at \url{https://github.com/sb895/chemical-exposure-information-corpus}} datasets were used. The third dataset diagnosisCodes could not be obtained (neither from the authors of ML-NET nor of the original paper~\citep{diagnosisCode}). We aggregate sentence labels to the whole description for cancerHallmarks and chemicalExposure, as was done for ML-NET.

\begin{table}[h!]
\caption{Multilabel classification performance@0.05 on a single run.}
\label{table:additionalResults}
\vspace{2mm}
\begin{subtable}[t]{.95\columnwidth}
  \caption{DistilBERT (NLP) + classification head on cancerHallmarks.}
  \label{tab:cancerHallmarks}
\centering
\begin{tabular}{l rrrrr}
\toprule
Loss  & \rotatebox{0}{weightedF1} & \rotatebox{0}{microF1} & \rotatebox{0}{macroF1} & \rotatebox{0}{Precision}\\
\midrule
$\mathcal{L}_{\text {BCE}}$ & 0.0 & 0.0 & 0.0 & 0.0 &\\
$\mathcal{L}_{\text {FL}}$ & 10.8 & 19.0 & 4.4 & 7.1 &\\
$\mathcal{L}_{\overline{\mathit{F1}}}$ & 17.0 & 17.6 & \textbf{9.8} & \textbf{8.9}\\
$\mathcal{L}_{\widetilde{\mathit{F1}}}$ & \textbf{20.2} & \textbf{31.3} & 9.5 & 5.9\\
\bottomrule
\end{tabular}
\end{subtable}

\vspace{2mm}
\begin{subtable}[t]{.95\columnwidth}
  \caption{DistilBERT (NLP) + classification head on chemicalExposure.}
  \label{tab:chemicalExposure}
\centering
\begin{tabular}{l rrrrr}
  \toprule
  Loss  & \rotatebox{0}{weightedF1} & \rotatebox{0}{microF1} & \rotatebox{0}{macroF1} & \rotatebox{0}{Precision}\\
                                                                                                                      \midrule
$\mathcal{L}_{\text {BCE}}$ & 5.1 & 5.8 & 1.2 & 4.7\\
$\mathcal{L}_{\text {FL}}$ & 26.8 & 34.8 & 9.3 & 13.0\\
$\mathcal{L}_{\overline{\mathit{F1}}}$ & 21.8 & 19.4 & \textbf{13.3} & \textbf{15.5}\\
$\mathcal{L}_{\widetilde{\mathit{F1}}}$ & \textbf{31.9} & \textbf{43.2} & 11.3 & 9.1\\
\bottomrule
\end{tabular}
\end{subtable}
\end{table}

For arXiv2020, moviePosters, cancerHallmarks and chemicalExposure, we saw after a few preparatory training rounds that almost only \emph{sigmoidF1} had non-zero results for \(t = 0.5\). Class representation is a lot more sparse for these dataset, we thus set the evaluation metrics threshold to a reasonable value of $0.05$ and train for $100$ (arXiv2020, moviePosters) or $500$ (cancerHallmarks and chemicalExposure) epochs until reaching convergence. Once thresholds were decided upon, no further threshold-hacking was performed. Note that a threshold of 0.8 on Pascal-VOC, as used by~\citet{ASL}, does not alter the results.

On the smaller chemicalExposure and cancerHallmarks datasets (see Tables~\ref{tab:cancerHallmarks} and~\ref{tab:chemicalExposure} respectively), the \emph{unboundedF1} loss delivers good results for macroF1 and Precision and the \emph{sigmoidF1} loss leads to higher scores on the remainder of the metrics. We observe that \emph{unboundedF1} scores higher than \emph{sigmoidF1} on macroF1 on the two small text datasets (chemicalExposure and cancerHallmarks). Since \emph{unboundedF1} forgoes thresholding altogether, we hypothesize that \emph{unboundedF1} develops tolerance for sparse datasets with low number of class instances.

\begin{table}[h!]
\caption{Descriptive statistics of all datasets.}
\label{table:allDbatasets}
\centering
\begin{tabular}{l l rrr}
\toprule
& Type & Classes & Average label count & Number of examples \\
\midrule
moviePosters & image & 28 & 2.2 & 37,632\\
arXiv2020 & text & 155 & 1.9 & 26,558\\
chemExposure & text & 38 & 6.1 & 3,661\\
  cancerHallmarks\hspace{-.7em}  & text & 33 & 3.5 & 1,582\\
  Pascal-VOC & image & 20 & 1.6 & 9,963\\
  MS-COCO & image & 80 & 2.9 & 122,218\\
\bottomrule
\end{tabular}
\end{table}

Notably for the cancerHallmarks dataset, predictions from a model trained with cross-entropy do not reach high enough values to surpass the threshold and thus all metrics return zero values. This was further observed during experimentation, thus cross-entropy loss might not be a good fit for solving small-dataset multilabel problems.

\end{appendices}

\end{document}